\documentclass[lettersize,journal]{IEEEtran}
\usepackage{amsmath,amsfonts}
\usepackage{algorithmic}
\usepackage{algorithm}
\usepackage{array}
\usepackage{color}
\usepackage[caption=false,font=normalsize,labelfont=sf,textfont=sf]{subfig}

\usepackage{longtable}
\usepackage{multirow}
\usepackage{epsfig} 
\usepackage{hyperref}

\usepackage{textcomp}
\usepackage{stfloats}
\usepackage{pifont}
\usepackage{url}
\usepackage{verbatim}
\usepackage{graphicx}
\usepackage{cite}
\def\BibTeX{{\rm B\kern-.05em{\sc i\kern-.025em b}\kern-.08em
    T\kern-.1667em\lower.7ex\hbox{E}\kern-.125emX}}
\usepackage{balance}

\begin{document} 

\title{Federated Learning in Intelligent Transportation Systems: Recent Applications and Open Problems}

\author{Shiying Zhang, Jun Li, Long Shi, Ming Ding, Dinh C. Nguyen,\\ Wuzheng Tan, Jian Weng, and Zhu Han
\thanks{
Shiying Zhang, Jun Li and Long Shi are with the School of Electrical and Optical Engineering, Nanjing
University of Science and Technology, Nanjing 210094, China (e-mail: \{shiying.zhang, jun.li\}
@njust.edu.cn and slong1007@gmail.com).

Ming Ding is with Data61, CSIRO, Eveleigh, NSW, Australia (e-mail:
Ming.Ding@data61.csiro.au).

Dinh C. Nguyen is with the Elmore Family School of Electrical and Computer Engineering, Purdue University, West Lafayette, IN, USA (e-mail: ncdinhbk@gmail.com).

Wuzheng Tan and Jian Weng are with the College of Cyber Security, Jinan University, Guangzhou 510632, China (e-mail: wzhengtan@163.com and cryptjweng@gmail.com).

Zhu Han is with the Electrical and Computer Engineering, University of Houston, Houston, Texas, USA (e-mail: hanzhu22@gmail.com).
}}



\maketitle

\begin{abstract}
Intelligent transportation systems (ITSs) have been fueled by the rapid development of communication technologies, sensor technologies, and the Internet of Things (IoT). Nonetheless, due to the dynamic characteristics of the vehicle networks, it is rather challenging to make timely and accurate decisions of vehicle behaviors. Moreover, in the presence of mobile wireless communications, the privacy and security of vehicle information are at constant risk. In this context, a new paradigm is urgently needed for various applications in dynamic vehicle environments. As a distributed machine learning technology, federated learning (FL) has received extensive attention due to its outstanding privacy protection properties and easy scalability.
We conduct a comprehensive survey of the latest developments in FL for ITS. Specifically, we initially research the prevalent challenges in ITS and elucidate the motivations for applying FL from various perspectives. Subsequently, we review existing deployments of FL in ITS across various scenarios, and discuss specific potential issues in object recognition, traffic management, and service providing scenarios. Furthermore, we conduct a further analysis of the new challenges introduced by FL deployment and the inherent limitations that FL alone cannot fully address, including uneven data distribution, limited storage and computing power, and potential privacy and security concerns. We then examine the existing collaborative technologies that can help mitigate these challenges. Lastly, we discuss the open challenges that remain to be addressed in applying FL in ITS and propose several future research directions.

\end{abstract}


\section{Introduction}\label{1}

\IEEEPARstart{D}{riven} by overwhelming demand for safe and reliable transportation systems, artificial intelligence (AI) has been wildly applied in the field of the Internet of Vehicles (IoV), called intelligent transportation systems (ITSs)\cite{Patel2019ASO}. 
The components comprising ITS encompass vehicle nodes, sensors, Roadside Units (RSUs), main base stations, and more. Notably, vehicles are equipped with on-board units that serve as network nodes, facilitating communication with both stationary and mobile RSUs, as well as other vehicle nodes. Furthermore, on-board sensors are employed to gather pertinent status information, while emergent messages generated thereby are transmitted to nearby vehicles and RSUs, subsequently relayed to the control center.
As the primary steadfast node within ITS, the roadside static RSU typically boasts dual communication interfaces, furnishing access points for vehicles. Additionally, the repertoire of roadside static facilities also encompasses wireless battery-powered sensor nodes. For disparate static sensor types, the integration of data from varied sensors frequently necessitates the utilization of data fusion techniques, encompassing infrared and camera systems.
In the realm of ITS, vehicular interconnectivity is presently realized through contemporary sensing mechanisms and Dedicated Short-Range Communications (DSRC), thereby facilitating route planning\cite{Abdelrahman2020CrowdsensingBasedPD}, traffic flow prediction\cite{Chen2021TrafficFP}, object detection\cite{Shi2021OrientationAwareVD}, infotainment broadcasting, and the provision of public services, all complemented by the integration of AI\cite{deOa2021ServiceQS}. However, the limited communication range inherent in DSRC technology mandates the deployment of dense fixed RSUs to ensure sufficient coverage, albeit this approach of extensively deploying static facilities entails exorbitant costs. Moreover, the application of AI technology tends to introduce time-consuming and computationally intensive challenges, rendering it more suitable for scenarios exhibiting lower sensitivity to latency.
While in complex and dynamic IoV networks, it is necessary to take into account low transmission latency and massive data storage, while promoting the information accuracy. Vehicles in the network must perceive the environment in real-time and make optimal decisions in the real-world. Consequently, any decision-making errors in the system, which may come from excessive network latency or malicious third-party attacks, unavoidably cause serious traffic accidents and irreparable loss of life and property.

In the following, we briefly summarize main characteristics of ITS and urgent issues in ITS:

\subsubsection{\textbf{High Response Latency}}

The high dynamic nature of vehicles places extremely high real-time requirements for their decision-making, and excessive response latency can pose serious safety risks. However, traditional IoV often experiences high response latency due to factors such as heterogeneous storage and computing resources, and unstable communication links.
\begin{itemize}
\item  \textit{Unstable communication links:}
In an IoV system, despite static and stable roadside infrastructure, the vehicles within the jurisdiction of the Roadside Units (RSUs) are constantly changing due to their own movement. Such frequent changes in network topologies will render communication links unstable, and disrupt route establishment. From the perspective of optimizing communication links, a number of existing works propose multi-hop routing path models\cite{Deng2020AMV}, cache-aided relays, as well as other methods\cite{Xia2020CacheaidedME}. However, these techniques can only be used to mitigate the issue of link failure, and the alternating process of routing will boost the latency.

\item  \textit{Data multi-modality and resource heterogeneity:}
Vehicles are often equipped with various types of sensors for data collection, access control, and action feedback. 
These sensors, roadside detectors, and other devices are distributed throughout the ITS data collection layer to collect relevant data, including vehicle speed, traffic flow density, vehicle location, and more\cite{Zhu2019BigDA}. To compensate for the lack of a single data model, sensor data regularly contains multiple modes, including one-dimensional signal data in smart grids, and two-dimensional image data for license plate recognition. Accordingly, the storage and computing capacities of distinct vehicles are limited and unevenly distributed. Specifically, different vehicles have distinct resource constraints. The immediate problem caused by this multi-modality and heterogeneity lies in that the duration of training for each vehicle varies greatly, resulting in the ``Cannikin Law" in system-level decision making. Despite extensive works on Mobile Edge Computing (MEC) resource allocation, frequent interactions between terminals and servers can also cause significant channel congestion.
\end{itemize}

\subsubsection{\textbf{Privacy and Security Issues}}
In the realm of IoV, the exchange of information relies on wireless communication links such as vehicle-to-vehicle, vehicle-to-RSU, and RSU-to-server connections. However, the high level of connectivity across different layers within ITS renders it susceptible to external attacks during model training, both in terms of data and network security.
For instance, an attacker can disrupt a vehicle's operations and provide misleading instructions, potentially leading to severe traffic accidents\cite{Lim2020FederatedLI}. Moreover, the presence of malicious vehicles within the ITS ecosystem can collude to manipulate or steal sensitive data, including vehicle license plate details, driver information, and location data. Given the low-latency requirements of ITS, the data security processing cycle is limited, thus necessitating robust data privacy protection mechanisms.
Furthermore, as technologies like Software Defined Networking (SDN) and fog computing continue to advance, network security emerges as a critical concern in ITS. While some existing research explores network security solutions in the context of IoV, these often rely on access control at the edge nodes and protection through predefined rules, such as firewalls, intrusion detection systems, and authentication-based approaches, which have high complexity and poor scalability\cite{Hamida2015SecurityOC}. However, there is a need for a comprehensive survey of novel architectures that address these challenges more comprehensively.

\subsubsection{\textbf{Low Decision-making Accuracy}}
The predominant reasons for the low accuracy of the model are data island and immature data fusion technology on the basis of dynamic decision-making.

\begin{itemize}
\item  \textit{Data island:}
Vehicle data privacy requirements lead to data island, where sensitive data is not allowed to be exchanged directly between vehicles. In addition to the vehicles themselves, data interaction between RSUs that control different vehicles is also prohibited. This places restrictive requirements on the data collected by the vehicle's sensors. Nonetheless, on the one hand, limited by the number of vehicle sensors and the frequency of real-time data collection, less information satisfies the training requirements, especially the label data. On the other hand, rare events including severe weather and holidays are sparse, particularly in certain hazardous scenarios. Real-world data collection is extremely difficult, which results in poor generalization ability of the model. Although some works can predict the flow in extreme states based on observing the regular state\cite{Yu2017DeepLA} or by utilizing the game engine to generate virtual datasets for model training\cite{Yue2018ALP},\cite{Gaidon2016VirtualWorldsAP}, the effect remains insignificant. In addition to the development of algorithms and original data, we intend to investigate the solution of this problem from a variety of novel perspectives.

\item  \textit{Imperfect sensor data fusion for dynamic decision-making:}
Since the expected performance is challenging to achieve with the aid of a single sensor, numerous solutions use multi-sensor fusion techniques to collect the samples\cite{Ding2020LongitudinalVS},\cite{Yang2020MultiSensorM}. In ITS, the sensor system employed by the vehicle includes multiple sets of cameras, radar, GPS systems, and inertial measurement units (IMUs)\cite{Kiran2022DeepRL} for environmental perception and vehicle positioning. Environmental perception is typically used by vehicles to detect their surroundings, such as road, pedestrian, and traffic light detection, while vehicle positioning uses GPS data to control vehicles and provide public services. Although researchers have proposed solutions on the basis of multi-sensor and multi-modality\cite{Manjunath2018RadarBO},\cite{Grigorescu2019NeuroTrajectoryAN}, the feasibility of the current multi-sensor fusion technology is still not quite enough, and real-time fusion has stringent demands on computing and communication performance.
\end{itemize}

Therefore, at this stage, the development of a sufficiently reliable ITS system remains a daunting challenge. 
Recently, a novel architecture deployed in ITS is widely investigated to meet the needs of timeliness, accuracy, as well as robustness in the ITS network.
In 2016, the Google team proposed a Federated Learning (FL) algorithm framework for the privacy protection of mobile Internet mobile terminals, and it has been utilized extensively in the medical, industrial, academic circles and other fields. 
For example, the recent GenoMed4All project led by the European Union aims to connect European clinical databases of rare diseases through FL platforms and establish supporting international datasets as well as interoperability standards. Reina \emph{et al.}\cite{Reina2021OpenFLAO} use the OpenFL architecture for the first time to train ML models in an international consortium of healthcare organizations. In addition, open source frameworks include FedML, UniFed, and others\cite{Liu2022UniFedAB}.
FL can use distributed samples residing on edge devices to collectively train a global model while observing data privacy regulations. This process is coordinated by a trusted third-party central entity, ensuring that participating entities do not directly interact with the data. Through continuous iterations, the central entity averages the local models and arrives at the optimal global model, which is subsequently distributed to the participants.

Although the central cloud-based architecture can process and make decisions based on global information, collecting data from distributed agents results in high bandwidth consumption and latency. At the same time, the decisions of local agents are limited by the local environment and cannot reflect the global characteristics. In contrast, the FL-enabled architecture within an ITS framework facilitates the deployment of large-scale caching and computation offloading mechanisms. This enables the training of a robust global model while preserving the privacy of vehicles and users. Furthermore, the adoption of vehicle selective aggregation and asynchronous aggregation algorithms allows vehicles that meet the predetermined threshold conditions in uploading local models and downloading global models. Consequently, this approach effectively mitigates the information imbalance problem.

In summary, FL is expected to accomplish the following key advantages:

\begin{itemize}

\item \textbf{Lower Communication Costs:}
Compared with centralized systems that directly aggregate large amounts of data to the server, FL only shares model parameters between the server and the clients\cite{Ma2020WhenFL}. Furthermore, for real-time decision-making scenarios, FL can be processed directly locally at the edge node, which is much more sensitive than that of convergence to the cloud for unified decision-making. Additionally, FL can allocate clients' weights in accordance with distinct bandwidths to increase communication effectiveness and reduce transmission latency\cite{Deng2022LowlatencyFL}.

\item \textbf{Privacy Protection:}
FL imposes constraints on the direct access of data in conventional centralized learning. It leverages locally trained models derived from edge-side data to make informed decisions. The data remains securely stored on local devices, with no inter-client sharing. Moreover, the training process of models on end devices predominantly relies on shared computational resources rather than centralized servers, thereby affording substantial safeguards for data privacy\cite{Wei2023PersonalizedFL}. Furthermore, FL can be synergistically integrated with other privacy and security protection mechanisms, rendering it adaptable to diverse ITS scenarios. For instance, the incorporation of decentralized blockchain facilitates the sharing of information among vehicles, emancipated from the jurisdiction of central server governance. Similarly, the fusion of FL with selective security aggregation, in tandem with MEC, fortifies privacy safeguards. Additionally, the adoption of proactive defense mechanisms such as differential privacy (DP) facilitates the generation of shared models embellished with noise\cite{Mothukuri2021ASO}.

\item \textbf{Scalability:}
In FL, the clients are coordinated by the server, which can flexibly allocate their participation level. Meanwhile, it allows the model to continuous train and make real-time adjustments on each participant's local device. This implies that the model can promptly adapt to newly collected data from in-vehicle sensors within a highly dynamic ITS environment. The architecture is especially advantageous for dynamically switching vehicles and edge servers participating in ITS. Additionally, this excellent scalability allows multiple data holders to participate in FL training\cite{Khan2021FederatedLF}.

\item \textbf{Robustness:}
As a distributed architecture, the performance of the final global model is jointly determined by multiple local models, resulting in a more stable FL system. Vehicles in ITS frequently switch between RSUs, meanwhile, the instability of the communication link can inevitably result in node loss. However, the characteristics of FL can bring it into play against this effect.

\item \textbf{Pre-conditions for Non-uniform Dataset:} 
Conventional distributed machine learning algorithms commonly assume a balance in node loads, which is often impractical in the context of ITS due to variations in device performance, uneven distribution of vehicles, and dissimilarities in data collection capabilities. In contrast, FL acknowledges this limitation and numerous research studies address this imbalance by formulating diverse strategies for model updates, aggregation, and numerical optimization.
Prominent approaches encompass grouped FL aggregation methods\cite{Chen2020SemiFederatedL}, adaptive weight update methods\cite{Kopparapu2020FedCDIP}, and the realization of personalized model representations through multi-model optimization algorithms\cite{Liang2020ThinkLA}. Furthermore, numerical optimization directly optimizes the data itself, thereby ensuring that the data distribution among clients aligns as closely as possible with the overall data distribution\cite{Zhao2018FederatedLW}. These well-established techniques can readily be applied to the ITS domain.

\item \textbf{Support Local Personalized Decision:}
Since FL is trained on the basis of distributed user data, each vehicle performs vertical FL or FL-based model transfer by collecting data with the same label but distinct features, thus supporting autonomous decision-making during local training.
\end{itemize}

The applications of FL in ITS systems are advancing with the emergence of privacy-enhanced frameworks\cite{Smith2017FederatedML}, through which vehicles can share knowledge without compromising the local context, and use the information continuously captured by sensors locally to enhance model performance.

The FL-enabled architecture within an ITS framework facilitates the deployment of large-scale caching and computation offloading mechanisms. This enables the training of a robust global model while preserving the privacy of vehicles and users.
There has been a flood of academic research on the applications of FL in ITS, including the challenges of FL deployment (e.g., model security, clients selection and scheduling, communication link stability, etc.), research on FL-enabled ITS scenarios (e.g., human monitoring, object detection, vehicle trajectory prediction, etc.). For example, oneVFC, a platform based on vehicular fog computing built by Phung \emph{et al.}\cite{Phung2021oneVFCAVF}, can coordinate information flow and computation tasks on vehicular fog nodes by managing distributed resources, which significantly reduces the program processing time in a real-life system. For integrating vehicle-to-everything communication with heterogeneous computation power aware learning platform, Pervej \emph{et al.}\cite{Pervej2022MobilityCA} propose a mobile-aware online FL-based platform for providing near-ground multilevel speed and vehicle-specific power prediction. Wang \emph{et al.} \cite{Wang2020AEBISAB} propose the electric vehicle integration system called AEBIS. The smart grid platform deploys FL in a network consisting of distributed generation units, power consumers and storage systems for power management.

\subsection{Current State of Art and Our Contributions}
In this subsection, we conduct a comprehensive survey of recent literature reviews on FL, blockchain, deep learning, as well as their integration within ITS. By combing through these reviews, we aim to provide a comprehensive understanding of the current state of research and identify potential research gaps and opportunities.

The study conducted by Manias \emph{et al.}\cite{Manias2021MakingAC} explore the challenges of machine learning-based ITS, focusing on dynamic properties such as dynamic environments, privacy, and data storage. The authors also discuss the rationale behind adopting FL architectures in ITS, highlighting the benefits of distributed and continuous collaborative learning.
In a study by Tan \emph{et al.}\cite{Tan2020FederatedML}, two applications of FL in ITS are cited: resource management and performance optimization. They briefly summarize the current issues related to device interference, data security, and the Non-independent-and-identically-distributed (Non-IID) of data.
Du \emph{et al.}\cite{Du2020FederatedLF} examine the application of FL in wireless Internet of Things (IoT), dividing IoT into sensing layer, network layer, and application layer, moreover, they propose several challenges and directions for integrating FL and IoT architectures.
Billah \emph{et al.}\cite{Billah2022ASL} investigate the application of FL and blockchain in ITS, particularly addressing privacy concerns and proposing a preparation approach for blockchain-supported FL.
Jamil \emph{et al.}\cite{Jamil2022ACS} explore the application of digital twin (DT) and FL in various contexts such as industrial IoT (IIoT), IoV, and the Internet of Drones (IoD).
Under the topic of IoV security and privacy issues, Hussain \emph{et al.}\cite{Hussain2022CyberSA} discuss cybersecurity and privacy issues in Connected and Autonomous Vehicles (CAVs) under an FL architecture, while Xing \emph{et al.}\cite{Xing2023ASO} elaborate on the attack model of the Social Internet of Vehicles (SloV) and provide a comparative analysis of typical data security solutions, emphasizing the synergy of federated learning in data security protection.
Other studies focus on the integration of blockchain and FL in ITS. Zhu \emph{et al.}\cite{Zhu2022BlockchainEnabledFL} discuss UAV edge network architectures that support blockchain and FL. Javed \emph{et al.}\cite{Javed2022IntegrationOB} discuss computational cost, communication overhead, and privacy issues in ITS, which can be addressed through the integration of blockchain and FL.
Recently, Chellapandi \emph{et al.}\cite{Chellapandi2023ASO} have compiled recent advances in applying FL in vehicular networking, specifically covering tasks such as object perception, motion planning, and vehicle control.

While these studies provide valuable insights into FL-enabled ITS, they also have their limitations. For instance, Manias \emph{et al.}'s work \cite{Manias2021MakingAC} primarily focuses on system robustness rather than specifically highlighting the advantages of FL itself. Tan \emph{et al.}\cite{Tan2020FederatedML} omit mentioning common scenarios in the application section, except for resource management and optimization. Similarly, studies by Zhu \emph{et al.}\cite{Zhu2022BlockchainEnabledFL} and Billah \emph{et al.}\cite{Billah2022ASL} concentrate on blockchain-assisted FL architectures and do not comprehensively cover other assistive technologies. The works by Hussain \emph{et al.}\cite{Hussain2022CyberSA} and Xing \emph{et al.}\cite{Xing2023ASO} solely discuss the advantages of FL in terms of privacy protection and security. Finally, with the exception of Billah \emph{et al.}\cite{Billah2022ASL} and Chellapandi \emph{et al.}\cite{Chellapandi2023ASO}, the other articles lack organization based on ITS task scenarios, which hinders the reader's intuitive understanding of FL-enabled ITS.

Table \ref{table1} summarized the differences between the existing works and this work. These limitations inspire us to do a more comprehensive research on FL-enabled ITS. On one hand, we aim to provide newcomers to this field with an intuitive and rapid understanding of the existing work through a detailed depiction of the system architecture and clear categorization of scenarios. Additionally, we strive to enable researchers in related areas to stay abreast of the latest research developments and identify future reference directions.
On the other hand, we have noticed that the majority of existing works primarily focuse on the privacy-preserving aspects of FL when applied to ITS, often neglecting other remarkable features of FL, such as extensibility and robustness. In our review, we intend to explore the potential value of FL and its integration with other technologies from a fresh perspective. This approach is essential for advancing both the theoretical research of FL and facilitating future ITS system upgrades and deployments.
In this paper, we first extensively discuss the state-of-the-art technical applications of FL in various ITS scenarios, and then analyze how the relevant FL auxiliary mechanisms solve a number of significant problems that have garnered considerable attention. Finally, we summarize the existing limitations of FL-enabled ITS and highlight future research directions. 
The following are the principal contributions of this paper:

\begin{itemize}
\item We sort out the issues existing in the current centralized ITS, in combination with the inherent advantages of FL, and analyze the motivation for employing the FL framework in ITS.

\item We gain our own insight by dividing ITS into four scenarios, including traffic management, object recognition, service providing, and traffic status identification, in accordance with the tasks in distinct stages (perception stage, prediction stage, decision stage). Subsequently, we separately describe the existing FL-enabled technologies and architectures in each scenario.

\item We provide a summary of the existing ITS's problems and challenges, including resource constraints, transmission constraints, security and privacy concerns, and uneven data and device distribution. Moreover, we organize some recent works and illustrate how FL solves the corresponding problems. And how to incorporate some supplementary mechanisms into the FL architecture to compensate for FL's deficiencies, which have never been systematized in any of the existing surveys as of this writing.

\item Eventually, we point out the current remaining open challenges and put forth future research directions for FL-enabled ITS.
\end{itemize}

\begin{table*}[!t]
\begin{center}   
\caption{Survey papers of FL-enabled ITS.} 
\renewcommand\arraystretch{1.2}
\begin{tabular}{|m{0.7cm}<{\centering}|c|p{1.4cm}|p{1.4cm}|p{4.6cm}|p{6.3cm}|}   
\hline     
\textbf{Ref.} & \textbf{Year} & \textbf{Framework} & \textbf{Scenarios} & \textbf{Main Topics}&
\textbf{Limitations}\\ 

\hline       \cite{Tan2020FederatedML}
&2020
&\ding{51}
&\ding{55}
& Resource management, performance optimization, and VN-based applications.
&The advantages of FL are not fully explained and the core ideas are on system robustness.
\\

\hline       \cite{Du2020FederatedLF}
& 2020
&\ding{51}
&\ding{55}
&Wireless communication and privacy.
&The authors do not elaborate on specific scenarios.\\

\hline       \cite{Billah2022ASL}
& 2022 
&\ding{55}
&\ding{51}
&Blockchain-enabled FL architecture, dataset and platform of each methods.
&The discussion of FL technology architecture is slightly lacking and covers incomplete scenarios.\\

\hline       \cite{Jamil2022ACS}
&  2022
&\ding{55}
&\ding{55}
& Digital Twins (DT) and FL in the Industrial Internet of Things (IIoT), the IoV and the Internet of Drones (IoD).
&The supporting literature cited by the authors for the application of FL in ITS is sparse and lacks focus.\\

\hline       \cite{Hussain2022CyberSA}
&  2022
&\ding{51}
&\ding{55}
& Cybersecurity and privacy issues for CAV.
&The discussion is limited to cybersecurity and privacy issues in CAV under FL architecture.\\

\hline       \cite{Zhu2022BlockchainEnabledFL}
&  2022
&\ding{51}
&\ding{55}
& An edge computing network for UAV with integrated blockchain and FL.
&The discussion is limited to the application scenarios of UAV under FL architecture.\\

\hline       \cite{Javed2022IntegrationOB}
&  2022
&\ding{55}
&\ding{55}
& Computational costs, communication overheads, and privacy issues.
& Only the role of blockchain as a supporting mechanism is discussed.\\

\hline       \cite{Xing2023ASO}
&  2023
&\ding{51}
&\ding{55}
& Social Internet of Vehicles (SloV), data security solutions, FL synergistic.
&The discussion is limited to the synergistic role of FL in data security protection.\\

\hline       \cite{Chellapandi2023ASO}
&  2023
&\ding{51}
&\ding{51}
& FL-enabled ITS, some application scenarios, resurce limitation, imperfect methodology, inadequate evaluation criteria.
&The scenarios involved are not comprehensive and do not explore the technical connections between the scenarios, and the transmission and resource inequality issues that most reflect the characteristics of ITS are not discussed.\\

\hline       \emph{Our paper}
& 2023 
&\ding{51}
&Object recognition, traffic status identification, traffic management, service providing.
& \multicolumn{2}{p{10.9cm}|}{
An extensive survey of FL-enabled ITS. Particularly,
\begin{itemize}
\item We provide an extensive explanation for the justification of implementing FL in ITS and propose a generic architecture for FL-enabled ITS.
\item To ensure logical clarity, we have introduced a novel categorization of four scenario types based on task phase and characteristics. Each scenario encompasses multiple sub-scenarios.
\item Beginning with an analysis of ITS characteristics, we integrate the limitations of FL-enabled methods. Furthermore, we compile a comprehensive inventory of current challenges and existing solutions, culminating in a summary of future research directions.
\end{itemize}
}
\\

\hline   
\end{tabular}   
\end{center}  
\label{table1}
\end{table*}

\subsection{Structure of The Survey}
\begin{figure*}[!t]
\center{\includegraphics[width=0.95\textwidth]{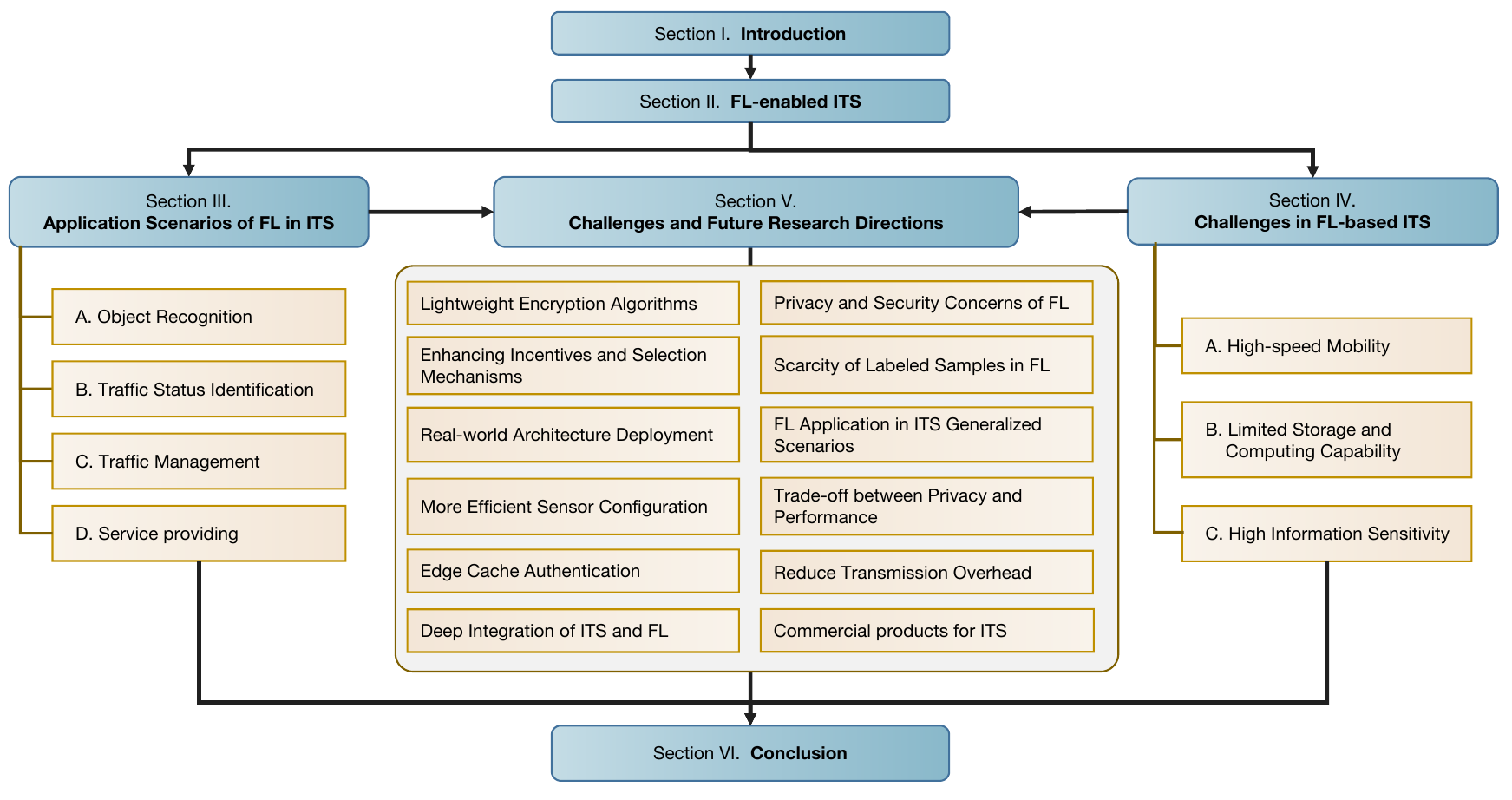}}
\caption{Structure of the survey.}
\label{figure1}
\end{figure*}

The remaining sections of this paper are structured as follows. Section \ref{2} presents several other reviews on the applications of FL to ITS, and we briefly describe how our work is distinct from that of others. In Section \ref{3}, we divide the applications of ITS into four scenarios on the basis of methods and task phases, and summarize the predominant problems and the application results of FL in each scenario. In Section \ref{4}, we categorize the challenges in ITS into three categories and organize further enhancements and optimizations based on FL architectures. In Section \ref{5}, we summarize several open challenges in the current ITS, and finally give a conclusion in Section \ref{6}.
Fig. \ref{figure1} outlines the organization of the survey.
Table \ref{table2} indicates the list of abbreviations employed in this paper.

\begin{table}[!t]
\begin{center}
\caption{Key Abbreviations.}
\begin{tabular}{ c | c }
\hline
 \textbf{Abbreviation} &  \textbf{Explanation} \\
\hline
ITS & Intelligent Transportation System\\
\hline
ADS & Autonomous Driving Systems\\
\hline
IoV & Internet of Vehicles\\
\hline
V2V & Vehicle-to-Vehicle\\
\hline
V2X & Vehicle-to-Everything\\
\hline
P2P & Point-to-Point \\
\hline
DSRC & Dedicated Short-Range Communications\\
\hline 
URLLC & ultra-reliable low-latency communication\\
\hline
SV & Smart Vehicle\\
\hline
EV & Electric Vehicle\\
\hline
CS & Charging Station\\
\hline
CSP & Charging Station Provider\\
\hline
VSP & Vehicular Service Provider\\
\hline
LPD & License Plate Detection\\
\hline
LPR & License Plate Recognition\\
\hline
QoI & Quality of Information\\
\hline
QoS & Quality of Service\\
\hline
RSU & Roadside Unit\\
\hline
MEC & Mobile Edge Computing\\
\hline
UAV & Unmanned Aerial Vehicle\\
\hline
FL & Federated Learning\\
\hline
RL & Reinforcement Learning\\
\hline
DRL & Deep Reinforcement Learning\\
\hline
Non-IID & Non-independent-and-identically-distributed\\
\hline
DP & Differential privacy\\
\hline
SGD & Stochastic gradient descent\\
\hline
HE & Homomorphic encryption\\
\hline
DNN & Deep Neural Network\\
\hline
CNN & Convolutional Neural Network\\
\hline
R-CNN & Region based-Convolutional Neural Network\\
\hline
GNN & Graph Neural Network\\ 
\hline
GCN & Graph Convolutional Network\\
\hline
GRU & Gate Recurrent Unit\\
\hline
LSTM & Long Short-Term Memory\\
\hline
SDN & Software Defined Networking\\
\hline
GPA & Global Passive Adversary\\
\hline
\end{tabular}
\end{center}
\label{table2}
\end{table}

\section{FL-enabled ITS}\label{2}

\begin{figure}[!t]
\center{\includegraphics[width=0.5\textwidth]{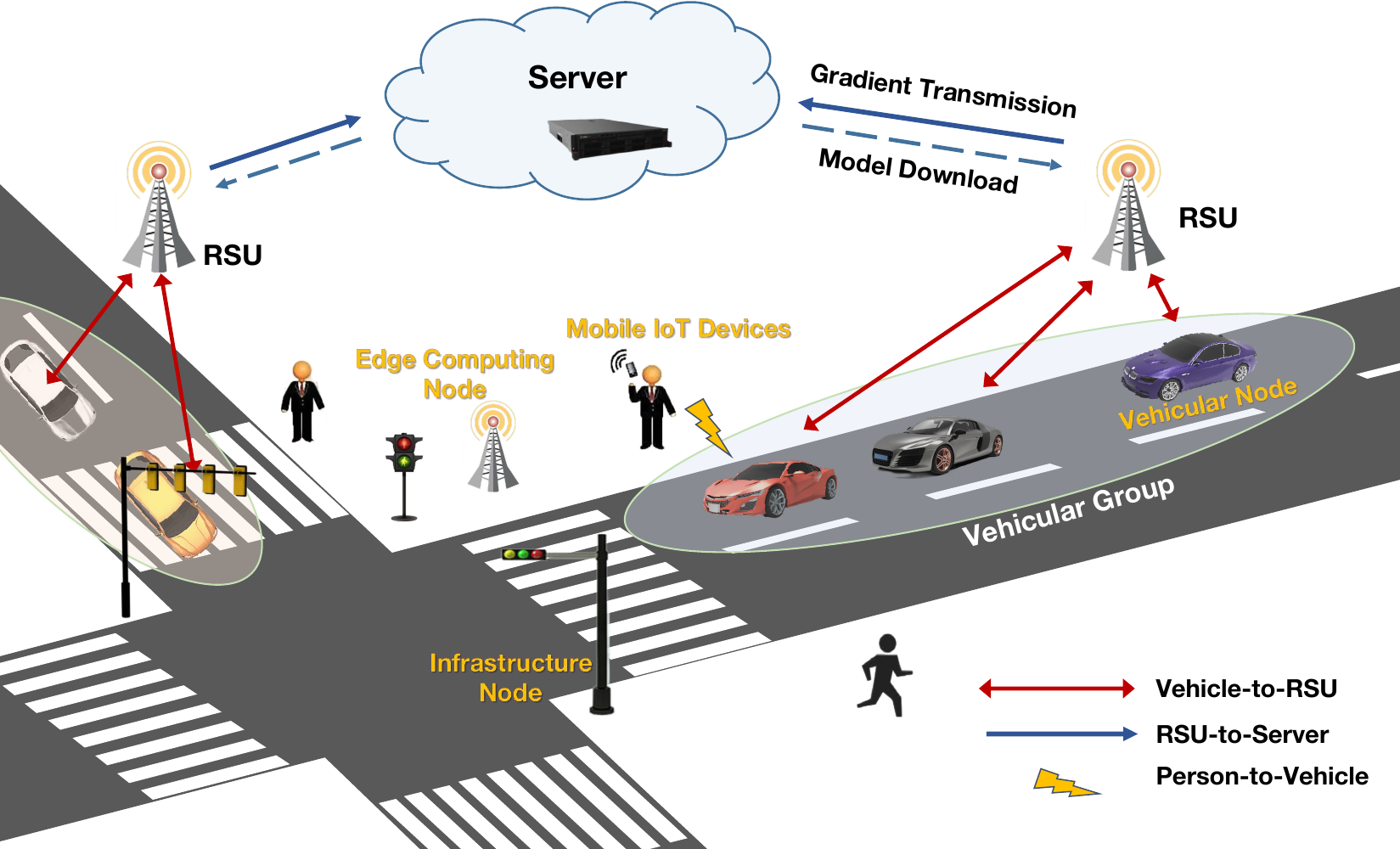}}
\caption{The architecture of FL-enabled ITS.}
\label{figure2}
\end{figure}

In the conventional centralized ITS, after data collection, vehicles and mobile devices are dispersed and transmit data directly to RSUs or the central cloud for training. To obtain higher model accuracy, a large number of high-quality images require to be transmitted, which occupies a large amount of bandwidth. In addition, wireless transmission makes it simple to receive external attacks FL supports local training for each vehicle, and the server is only used to aggregate model parameters and broadcast.

The ensuing exposition expounds upon the foundational architecture for integrating FL within the context of an ITS. 
Initially, the central cloud server direct oversight of vehicles throughout the entire area, each vehicle establishes an exclusive point-to-point (P2P) communication link with the cloud server, resulting in significant communication overhead. In order to accommodate the forthcoming communication networks characterized by high density and short-range connectivity, some researchers have embraced a hierarchical FL architecture, as illustrated in Fig. \ref{figure2}. This framework consists of three distinct layers: the top layer, the middle layer, and the bottom layer.

\textbf{Top Layer:} The top layer is the central server, the coordinator of the entire ITS. Its role encompasses the regulation of update frequency and participant count, as well as performing global caching and aggregation. In FL, the crux lies in the artful design of an aggregation scheme, which largely determines the performance of the global model. The conventional approach, known as FedAvg, involves the computation of model averages from the uploaded counterparts. Subsequently, numerous works have developed more refined aggregation algorithms to capture valuable information.

\begin{figure*}[!t]
\center{\includegraphics[width=0.88\textwidth]{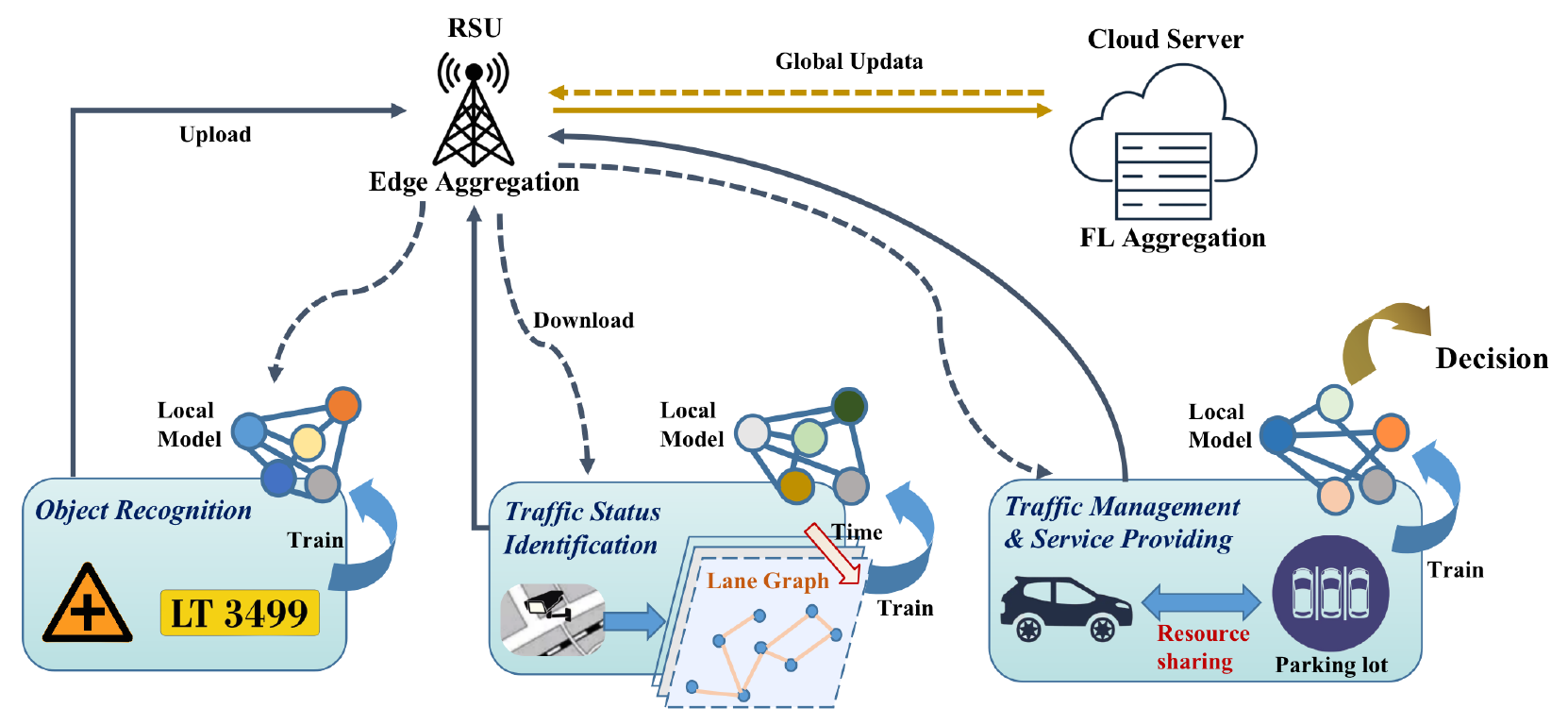}}
\caption{FL-enabled ITS in different scenarios.}
\label{figure3}
\end{figure*}

\textbf{Middle Layer:} The middle layer consists of the relay devices including RSUs, which are charged with acting as an intermediary for mobile devices including vehicles to interact with the server. Unlike the main server, RSUs have limited computing and storage capabilities. In addition to their capacity for aggregating and storing model parameters, RSUs possess the capability to retain a limited quantity of contextual information, including vehicle IDs, vehicle data, and the location particulars of the RSU itself, serving management objectives.
 Previous research has explored the use of infrastructure for deploying an edge control plane, where selected RSUs are designated as controllers, endowed with control functions within the edge network. This configuration enables them to act as edge controllers, enhancing adaptability to variations in the topology of the vehicular network.
During system operation, RSUs upload aggregation parameters to the central server. Subsequently, upon completing an aggregation round, the server dispatches the updated model parameters to the RSUs, and are then disseminated to each vehicle through a broadcasting mechanism.

\textbf{Bottom Layer:} The bottom layer, as the data collection layer, includes devices such as vehicles and roadside sensors that are tasked with collecting environmental data. This entails the employment of speed sensors, airbag sensors, and night vision sensors to capture vehicle safety data, as well as position sensors to provide real-time vehicle status parameters. These data collectors locally cache the real-time data within the ITS for training purposes, subsequently transmitting the trained local models to the RSUs within a designated timeframe. In instances requiring privacy safeguards, the trained models are encrypted to prevent malicious third-party agents from tampering with privacy. Nonetheless, owing to the limited computational capabilities of vehicles, the training may occasionally exceed the allocated timeframe due to excessive computing loads. In such circumstances, a flexible model upload time limit may be placed within the framework, or vehicles with limited computing capability can offload computing tasks to trusted RSUs or other edge servers to assist in model training and data storage.

In essence, vehicular nodes, edge computing devices, mobile IoT devices, and infrastructure nodes can all be regarded as clients in FL-enabled ITS. Vehicular nodes collect local sensor data and vehicle travel data, leveraging this information to conduct local model training and updates. Edge computing devices, typically located in infrastructure or cloud-edge servers near vehicles, play a coordinating role in facilitating communication and model aggregation among vehicular nodes. Mobile IoT devices, such as smartphones or in-vehicle terminals, can also serve as clients for FL, collecting vehicle data and communicating with other clients. Meanwhile, infrastructure nodes, such as traffic management centers or roadside devices, gather data from vehicles and other devices and use it for local model training and collaborative learning processes.

\section{Application scenarios of FL in ITS} \label{3}
We research and organize existing works by service types and scenarios, including such license plate recognition, autonomous controller design, congestion monitoring, knowledge sharing, route planning, etc., and divide them into four major scenarios to classify the specific scenarios, which include traffic management, object recognition, service provisioning, and traffic status identification.

The general basis of scene division is according to the distinct task phases of ITS. Fig. \ref{figure3} depicts a variety of FL application scenarios in ITS. The scene on the left is object recognition, including license plate recognition, traffic sign recognition, etc., the majority of which are image recognition-based classification detection tasks. In the middle is the traffic status recognition scenario, which is mostly on the basis of the analysis of time series data, including travel mode recognition and traffic flow prediction. The final phase includes vehicle management and service delivery. In this figure, the parking control is represented by P2P information sharing with the parking control center on the condition that the vehicle is in the state of finding a parking space, and decisions are made after completing the training.
In the following, we review and investigate the integration of FL and such ITS application scenarios with four different aspects of object recognition, traffic status identification, traffic management, and service providing.

\subsection{Object Recognition}
In ITS, the safety of Autonomous Driving Systems (ADS) is ensured by learning data captured by cameras and sensors in wireless networks. Object recognition is a basic work in the IoV\cite{fWu2021FedSCRSC}, principally in the dynamic system to process image tasks, owing to the extensive use of computer vision technology, the performance of the traditional object recognition model has been enhanced. Currently, there are a variety of algorithms for video frame processing, and they have found widespread application in vehicle tracking\cite{Gao2020ManifoldSN} and obstacle recognition\cite{Song2018RealTimeOD}, including region-based Convolutional Neural Networks (R-CNN) \cite{Girshick2014RichFH} and YOLO-based detection methods\cite{Redmon2016YouOL}. Typically, bounding boxes are drawn around road detection targets and sent to vehicles for route planning. Among them, the YOLO method is a more developed method currently, and the bounding boxes can be directly exported from the CNN model. Nonetheless, two-stage object detectors such as R-CNN use a more fine-grained paradigm to achieve higher detection accuracy when performing object recognition, which results in a longer detection time and increased complexity. While using a faster, more optimized one-stage object detector, it will lose model performance. At present, some works have been done to find a trade-off between the accuracy of the model and the speed of inference\cite{Huang2017SpeedAccuracyTF}. On other hand, the work in \cite{Oh2017ObjectDA},\cite{Du2017CarDF} and\cite{Bijelic2019SeeingTF} utilizing the method of multi-modal image data fusion, attempt to improve accuracy and the issue of missing labels. During the same time, in the target recognition scene, there are often some image occlusion problems, relying on the perspective captured by the sensors of a single vehicle cannot include sufficient features, and multi-mode data set based on distinct driving conditions and object labels can substantially enhance the network's precision and resilience. Considering the processing scenarios of ITS distributed images, it is natural to consider FL to alleviate the aforementioned issues.

As the first application of FL in vehicle target recognition, the authors of\cite{Jallepalli2021FederatedLF} verify through simulation that the FL architecture can identify detection targets that cannot be recognized in local training of a single vehicle. For instance, they use YOLO to perform five epochs of detection on four clients. It demonstrates that some unrecognized trucks and pedestrians at the original image's boundary can be detected using FL, indicating that other users indirectly gain knowledge through weight aggregation. On the basis of this work, to achieve better training results in vehicle object recognition scenarios supported by 6G, the work of \cite{Zhou2021TwoLayerFL} deals with the hierarchical structure of FL in ITS. The authors propose a two-layer FL model based on CNNs (TFL-CNN) to perform hierarchical model selection at the edge and in the cloud. RSUs use metrics such as RSU distance from vehicles to estimate model quality in model aggregation, and subsequently multiple RSUs upload aggregated parameters to the cloud, unlike the RSU layer, whereas the cloud infers the model quality (i.e., whether it is at a busy intersection with a great deal of data) and its computing capability (i.e., whether it can process a large amount of data) on the basis of the location of RSUs.

In addition to obstacle recognition, another noteworthy application of image recognition in ITS is fault recognition, which is commonly employed in UAVs. For instance, in the field of electric power systems, many companies usually choose to use UAVs instead of traditional manual inspections due to their compact size and flexible maneuverability. In this context, Liu \emph{et al.}\cite{Liu2022ObjectDO} pioneer the application of FL to a power line image detection task based on the YOLO v5 model. Their experimental findings demonstrate that FL training involving multiple power companies could yield a model with higher detection accuracy compared to local models. Similarly, Fang \emph{et al.}\cite{Fang2021ResearchOU} utilize an enhanced lightweight YOLO v3 model specifically tailored to the UAV scenario for target recognition.
However, when employing the FL architecture for image detection, there is a scarcity of real-time samples available for UAVs. This is due to the inherent mobility of UAVs, which makes it challenging to add samples during online training. To address this limitation, Wu \emph{et al.}\cite{Wu2022CIOFLCI} develop a collaborative inference-based online FL system named ``CIOFL". This system leverages a large-scale high-quality model to label real-world images through multi-node collaborative inference, thereby expanding the training dataset.

Most object recognition tasks are aimed at avoiding collisions, including road detection, vehicle detection, pedestrian detection, etc, which aim to identify circular boundaries and areas where vehicles may travel, and are applied to lane keeping and vehicle departure control systems, which can be categorized into the identification of obstacles. Additionally, there is another scenario for the recognition of license plate numbers and traffic signs\cite{Muhammad2021DeepLF}, which have higher image quality expectations. For instance, the license plate number recognition task is typically separated into two stages. The first stage is license plate detection (LPD), which is adopted to preliminarily determine whether the image is displayed as a license plate, and the second stage is license plate recognition (LPR), which is employed to detect license plate characters after completing LPD. There are three common approaches to LPR tasks: 1) area-based methods, 2) color-based methods, and 3) pixel-to-pixel methods. The predominant bottleneck of several methods at this stage is the processing of multi-directional license plate detection, as well as dealing with motion-induced blurred images\cite{Xie2018ANC}.
In \cite{Kong2021AFL}, the authors redesign the LPR model that can be deployed on edge devices, and simultaneously considered the issues of photo blur and orientation anomalies caused by the dynamic properties of vehicles and sensors, and proposed a license plate tilt correction algorithm to enhance the robustness of the model.
Aiming to reduce the computational overhead,\cite{Xie2022EfficientFL} designs a spike neural network (SNN) encoding scheme based on neuron receptive fields, which has better noise immunity and accuracy than FL using CNN.

Furthermore, it is important to highlight that while FL effectively addresses the issue of data fragmentation, there still exists a challenge in terms of limited availability of locally labeled data. This limitation arises due to the cost associated with data labeling, as well as the aforementioned constraints on image quality. To tackle this issue, researchers have turned to the utilization of semi-supervised learning (SSL) methods, which have proven to be effective. Building upon this, novel frameworks known as semi-supervised FL (SSFL) have been developed.
For instance, Zhang \emph{et al.}\cite{Zhang2022RobustSF} devise a SSFL framework specifically tailored to the task of UAV image recognition. In their work, they incorporated a dynamic hyperparameter term and two active learning strategies, which collectively contribute to achieving high-quality performance in SSFL. This approach helps overcome the scarcity of labeled data and enhances the accuracy and effectiveness of the image recognition process.

In general, the integration of FL into the target recognition scenario of ITS facilitates the construction of comprehensive training models by harnessing distributed data obtained from vehicle networks and road sensors. This approach effectively addresses challenges such as visual obstructions resulting from individual sensors or vehicles, as well as the ambiguity in identifying targets due to the dynamic movement of vehicles. However, the development of an FL architecture capable of real-time analysis of extensive multimodal data remains an arduous endeavor\cite{Yaqoob2020AutonomousDC}. Alongside the amalgamation of data modalities from diverse edge devices, it is imperative to satisfy the stringent requirements of low latency. Furthermore, to optimize the utilization of collected multi-layered resources, it is essential to incorporate road characteristics when designing sensor layouts and establish sophisticated simulators that provide high-fidelity image rendering within complex recognition scenarios, thereby minimizing perceptual errors.

\subsection{Traffic Status Identification}
 Traffic status identification, including travel mode identification and traffic speed forecasting, is frequently optimized to enhance public services and observe atypical vehicle driving behavior, understand driving intent and avoid potential traffic accidents\cite{Ho2017WiSafeWP}.
 In comparison with the transfer and applications of image learning algorithms in target detection and recognition, recognition of traffic status requires more collaborative training of sensors and auxiliary systems, including speed sensors, acceleration sensors, and vehicle positioning systems.
 
 \subsubsection{Vehicle Position}
Vehicle positioning is a basis for applications including vehicle navigation and lane keeping. Global Navigation Satellite System (GNSS) and Inertia Navigation System (INS) are currently the most advanced positioning systems, since the other two positioning methods can achieve rapid update frequency and high short-term accuracy, this method is too costly and highly dependent on the GNSS base station to be used for ADS algorithm testing\cite{Groves2007PrinciplesOG}. Therefore, multi-system-based data fusion is a development direction of vehicle positioning tasks, and it is coordinated with onboard sensors and visual positioning systems. Some existing methods rely on sensor-rich vehicles (SRVs), and yet this approach requires high implementation costs and requires a large number of bandwidth resources.

The work in\cite{Kong2022FedVCPAF} first employs the FL architecture based on the cooperative perception of V2V vehicles to generate more training samples, aiming to protect privacy whereas cooperating with universal vehicles to maximize the role of SRV, and for GPS error correction, and for errors caused by individual differences, transfer learning methods are used. Additionally, the work in\cite{Xu2020AnIT} uses remote sensing data as an aid to increase the detection range of vehicle congestion, which is employed to solve the issue of insufficient coverage of traffic monitoring systems, and FL as a privacy protection method as well.

\subsubsection{Travel Mode Identification}
Accurate data on travel patterns can help governments and related businesses better comprehend user travel habits and develop more effective traffic control strategies. As an extension of vehicle positioning service, it predicts trajectories and recognizes patterns by extracting features from a large number of GPS users and vehicle trajectory data. In the earliest work, a GPS device is used to record the travel location features at preset time intervals, the long trajectory is first divided into multiple segments with distinct patterns employing a change point clustering segmentation scheme, and subsequently the statistical features of each segment are adopted as an input of classifier\cite{Zheng2008UnderstandingMB}. With the popularity of neural networks, researchers mapped GPS trajectories into images, and extracted features using CNN and other techniques until the emergence of networks including Long Short-Term Memory (LSTM), which can capture spatiotemporal correlations to obtain better classification results. Nonetheless, several obstacles remain in this scenario, including vehicle trajectory-based classification task, which relies too much on GPS data. It is difficult to track vehicles in the absence of GPS signals. While some indoor positioning technologies, such as Wi-Fi\cite{Vasisht2016DecimeterLevelLW}, can be leveraged, deploying a large number of sensors locally will incur huge storage and computing costs.
The work in\cite{Gao2021GlowIT}, it is proposed to transfer the smartphone's inertial data to the vehicle in order to determine the vehicle's position in real-time. After the vehicle enters the closed area, the position inference is done by using the FL updated model, and the GPS data is only adopted for training in the open environment. During the same time, the use of FL also effectively alleviates the heterogeneous problems caused by distinct driving and equipment hardware distinctions of users.

In the real-world, vehicle users generally do not attach labels to the driving mode by themselves, and general institutions do not have the privacy permission to collect user labels.
 In\cite{Zhu2022SemiSupervisedFL}, the authors assume that the cloud server side holds a small privacy permission data set with a drive mode tag, whereas the user's data set lacks a tag. They design a semi-supervised FL framework that first employs the pseudo-labeling method to label local data before grouping and aggregating according to the class distribution as a standard to solve the Non-IID problem. In\cite{Koetsier2021FederatedCD}, the authors optimize the one-class support vector machine with stochastic gradient descent (SGD), so that the algorithm supports sequential learning, it can be utilized in FL to detect abnormal vehicle trajectories at intersections, thereby reducing the burden of local data labels.

 Except for that, some works consider the optimization of FL aggregation methods and training parameters in trajectory prediction or travel mode identification tasks. In\cite{Nakanoya2021PersonalizedFL}, the authors make personalized FL updates for distinct parameters in a trajectory prediction scenario. They come to the conclusion that one way to effectively prevent over-fitting during training is to slow down the update speed of parameters that share common features. For instance, some of the shared kinetic data might have similar similarities.
 In\cite{Torino2020FederatedLF}, the authors demonstrate that the aggregation method and the FedAvg method that give greater weight to comparable observations have better model performance with small moving windows and multiple communication rounds, correspondingly. It is crucial to highlight, however, that in LSTM-based vehicle prediction, the trained models perform poorly in predicting the departure time of vehicles from one cell to the next, perhaps as a point for future research.
 The work in\cite{Zhang2021DistributedDM} develops a three-stage map fusion method, including the density-based spatial clustering of applications with noise, the score-based average, as well as the intersection over union-based box pruning, global map fusion achieved at the edge server using knowledge extraction. And the method of FL generates training labels for FL when data labels are not available.
 
\subsubsection{Traffic Flow Prediction}
Traffic flow prediction compared with vehicle positioning and travel mode recognition, has more time and space dependence. Distinct locations and distinct time in the same location have an effect on the predicted value, this highly dynamic task scenario frequently needs to combine more edge-end information, and for congested vehicles to assign a data set (due to the fact that traffic congestion state and traffic flow are strongly correlated). Furthermore, the location of the camera, speed, and even weather conditions can be included in the input data set for better estimation of traffic behavior\cite{Silva2020TowardsFL}. 
Recently, GNNs have been applied to traffic flow prediction and estimation scenarios, where coordinate information associated with nodes is used to assist in routing to avoid information diffusion\cite{Liang2022SpatialTemporalAI}. The monitoring stations in the real scenario are located on the nodes of the GNN, the connection lines between stations are edges, and the adjacency matrix is calculated from the distance between stations.
Similar to sensor data, the adjacency matrix contains feature information and associative user privacy.
In some works employing GNN or GCN, the design of the aggregation method for the adjacency matrix is considered on the condition that aggregating with FL, which enhances the scalability of the network o improve application to traffic speed and traffic prediction\cite{Zhang2021FASTGNNAT},\cite{Xia2022ShortTermTF}.
In \cite{Liu2020PrivacyPreservingTF}, the authors combine FL and Gate Recurrent Unit (GRU) and add the Joint-Announcement protocol to the aggregation mechanism. This protocol, which has been applied to large-scale distributed prediction and also captures the time-space correlation of traffic data flow, uses random subsampling for participants to reduce communication overhead. 
In addition to traffic forecasting, some research also applies FL to the forecasting of passenger flow on urban rail systems \cite{Shen2020ABB},\cite{Yuan2022FedTSELF}.

Due to the high cost, many urban arterial roads lack traffic detectors, so they are typically only installed on highways, which is not conducive to the prediction of the entire network. Through the collaboration of FL training, this problem can be mitigated\cite{Tedjopurnomo2022ASO}.

\subsection{Traffic Management}
Object detection and traffic state identification rely on cooperative perception of the vehicle's environment and are used to characterize the scene and predict vehicle behavior. In traffic management scenarios, intelligent decisions are made based on the previous two, including parking control in a parking lot and traffic light control at an intersection, which are occasionally treated as game problems in multi-institutional collaborative training.
\subsubsection{Parking Management}
The parking control system for parking space search is one of the most important ITS activities, particularly in large cities\cite{Lin2017ASO}. On average, vehicles remain stationary 90$\%$ of the time. Several works point out that the dispatch center is crucial\cite{Khalid2018TowardsAC} to calculate the optimal route and make parking recommendations. Its basic components are a driver request center and an intelligent parking assignment center that makes reservations and checks the availability of parking spaces through the sensors and applications\cite{Geng2012AN}. Nonetheless, in most scenarios, there are multiple service providers, and cloud storage is required to centralize information, including connecting multiple service providers to a public cloud, and users extract information from it\cite{Atif2016InternetOT}. In order to utilize the onboard resources of parked vehicles to complete a given workload, the authors in\cite{Huang2021FedParkingAF} propose an appropriate incentive based on FL to encourage vehicles to enter the parking space. First, the parking capacity constraints of multiple parking lots are obtained through LSTM, and subsequently, the interaction between the parking lot and the vehicle is obtained. It is characterized as a non-cooperative game, and theoretical analysis proves the unique existence of the Stackelberg equilibrium.

Another task scenario in parking management is parking trajectory planning. Unlike route planning, this scenario introduces constraints related to narrow spaces. Specifically, in vehicle navigation, polynomial-based path planning is commonly used to generate multiple alternative collision-free and smooth curves. However, due to limited space, parking lots often have multiple non-differentiable points. Additionally, given the complexity of the environment, obstacle avoidance constraints in parking are significantly more severe than in normal driving lanes, and sufficient mobility is required. Currently, neural network-based approaches are popular. For instance, Dolgov \emph{et al.}\cite{Dolgov2010PathPF} initially use the Hybrid A* algorithm to sample the control space and obtain smooth dynamic curves. Subsequently, Fassbender \emph{et al.} \cite{Fassbender2016MotionPF} propose two node expansion methods: one employs numerical optimization to solve boundary problems, and the other generates edges using a simulated controller to guide the vehicle towards the global reference path.

\subsubsection{Traffic Signal Control}
In the traffic signal control scenario, the optimization directions include network optimization, intersection optimization, roundabout optimization, and timing cycle optimization\cite{Shaikh2022ARO}. 
On the one hand, recently, advanced machine learning techniques, such as evolutionary algorithms and swarm intelligence algorithms, have been applied to tackle optimization problems characterized by nonlinear, continuous, and discrete factors. One such problem is traffic signal control. Notably, Li \emph{et al.}\cite{Li2019MultiobjectiveOP} undertake a comprehensive approach by integrating various metrics, including system throughput, latency, and intersection volume overflow. They use genetic algorithms to address the optimization problem, conducting their investigations over three different levels of vehicular network complexity. Similarly, Jia \emph{et al.}\cite{Jia2019MultiobjectiveOO} present a metaheuristic algorithm based on particle swarm optimization to tackle multi-objective optimization challenges specifically in the realm of signal timing schemes. However, this approach does not consider hybrid traffic flow control models and vehicle coordination. 

On another front, reinforcement Learning (RL) has gained popularity owing to its ability to bypass the need for excessively idealized assumptions and intricate mathematical derivations. 
 Initially, the control unit collects state information, including queue length, and vehicle position, and then executes actions on the basis of the policy obtained by the RL method, and finally, the agent receives rewards from it. This procedure reduces intersection traffic congestion. Nonetheless, the high joint action space dimension makes centralized RL infeasible for large-scale traffic signal control\cite{Chu2020MultiAgentDR}. Recent works have also introduced FL into traffic control scenarios, allowing agents to communicate remotely without routing and load model parameters when idle.The relevant simulation demonstrates that this method is conducive to the algorithm's rapid convergence\cite{Wang2020AdaptiveTS},\cite{Ye2021FedLightFR}. In the follow-up work, a cooperative optimization framework can be considered to separate the traffic network into individual components to better optimize the traffic network containing numerous intersections.

In the context of traffic management scenarios, the coordination among multiple organizations is often necessary, leading to frequent interactions and substantial communication overhead in model aggregation. Thus, ensuring the reliability of FL for model training and inference in large-scale communication networks becomes paramount. Currently, network edge caching is recognized as an effective approach to enhance Quality of Service (QoS) in wireless network access. For instance, Li \emph{et al.}\cite{Li2023AFL} propose an edge cooperative caching scheme based on federated deep reinforcement learning, where collaborative models are formulated as a Markov process, enabling dynamic and adaptive caching. However, the practical implementation of high-precision map caching remains limited, with current efforts primarily focused on the theoretical domain. In summary, optimizing caching strategies in dynamic network topologies and accommodating high-speed node mobility within FL-enabled ITS presents an ongoing challenge.

\subsection{Service Providing}
In this section, we discuss two widely employed vehicle public service scenarios, charging services for new energy vehicles as well as route planning services for vehicle navigation. Due to the need to consider complex constraints in the actual public service environment, a truly mature system has not yet emerged. Here we only do the latest applications of FL, and briefly describe the shortcomings of the existing methods.
\subsubsection{Charging Service}
The primary categories of vehicles that require charging services encompass both UAVs and electric vehicles (EVs). Notably, UAVs have exhibited a shift towards civilian applications in recent years, finding extensive usage in areas such as parcel delivery and traffic surveillance. Simultaneously, due to high energy efficiency and low emissions, EVs have become one of the sustainable solutions for retrofitting traditional transportation systems. And the general setting is that once a request from EVs is received, the charging station provider (CSP) will focus on providing energy to the charge stations (CSs) to meet the electrical needs of EVs. In the traditional centralized architecture, the charging time of each EVs is determined by the aggregator, which straightforwardly collects the charging demands of EVs, and subsequently solves the optimization problem to determine the charging speed. Even though the global system state is readily accessible, failure to resolve optimization issues can result in system crashes, so a system backup must be considered. Additionally, additional complexity arises on the condition that the control variables and constraints of EVs are increased\cite{Nimalsiri2020ASO}. Recent research has introduced hierarchical FL into this scenario, grouping EVs and managing each group via sub-aggregators. In addition, due to the complexity of billing, highly sensitive signal control, and state estimation, some existing privacy protection methods based on data tampering are inapplicable to EVs, which provides additional justification for the implementation of FL.

In\cite{Wu2022ElectricVC}, the authors combine FL with random forests and CNN for power load prediction.
To save communication overhead, the work in\cite{Saputra2019EnergyDP} adopts a clustering-based approach on the CS side to reduce the data set dimension. Consequently, the biased prediction can be minimized.
Nonetheless, the authors did not fully consider the distinct features of EVs and CSs.
In \cite{Wang2021ChargingSR}, the authors design a cross-platform FL mechanism, and introduce a recommendation model with cross features for EVs and CSs, selectively adopting hash and RSA encryption, which can increase the convergence rate of the model while maintaining privacy.

Comparable to parking management, scenarios with multiple agents can also be optimized by introducing economic-driven games.
In\cite{Saputra2020FederatedLM}, the authors utilize a multi-principal one-agent (MPOA) contract-based economic model to convert CSs utility maximization into a non-cooperative energy optimization issue of MPOA contract strategy. They demonstrate the existence of balanced contract solutions for all CSs and develop an iterative algorithm for obtaining balanced solutions.
Considering that FL typically disregards the uncertainty associated with energy demand forecasting, the authors in\cite{Thorgeirsson2021ProbabilisticPO} generalize FedAvg, added the probabilistic prediction algorithm, as well as verified the performance advantage over the deterministic prediction model through experiments.
Zou \emph{et al.}\cite{Zou2023WhenHF} introduce an approach for urban UAV charging services, combining Hierarchical Federated Learning (HFL) with LSTM and stochastic game theory. In the first stage, LSTM is employed to forecast energy demand data, enabling proactive prediction. Subsequently, in the second stage, an optimal energy dispatching strategy is determined through the utilization of Markov games, leveraging the predicted demand as a basis for decision-making.

Indeed, it is apparent that when it comes to designing FL architectures for charging services, there are additional considerations and constraints compared to general traffic control scenarios. At the operator level, the priority lies in maximizing operational efficiency and minimizing costs. Similarly, at the aggregator level, the focus is on maximizing the benefits of aggregation and reducing power supply costs. At the individual electric vehicle (EV) level, the prime objective is to minimize charging power loss and battery degradation.
Moreover, market fluctuations play a significant role in causing fluctuations in electricity demand, while the sensitivity of time-varying heat loads introduces uncertainties in the network\cite{Ghavami2016DecentralizedCO}. It is imperative to effectively address these external disturbances and enhance the robustness of the proposed solutions. 
Furthermore, while the widespread and dense deployment of fixed charging stations as public infrastructure, combined with the utilization of mobile charging stations as supplementary resources, serves as the fundamental solution to alleviate range anxiety, challenges related to cost control and maximizing the utilization of limited charging stations still persist.

\begin{figure*}[!t]
\center{\includegraphics[width=1\textwidth]{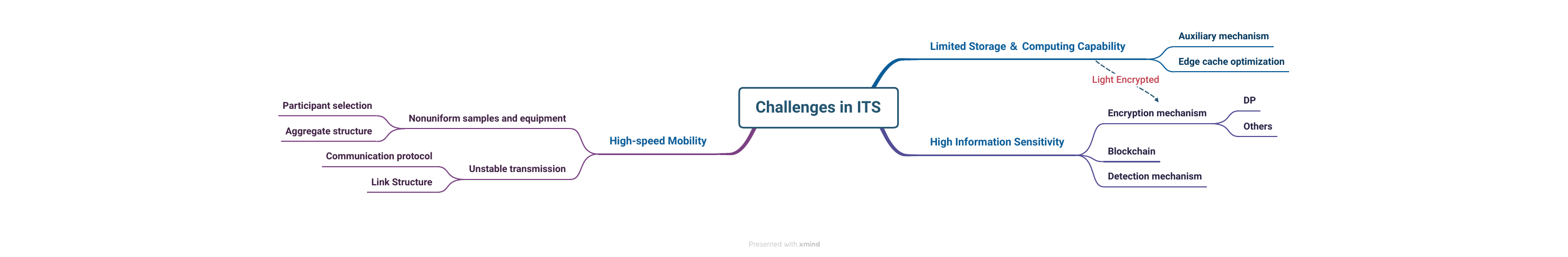}}
\caption{Challenges in ITS.}
\label{figure4}
\end{figure*}

\subsubsection{Route Planning}
The distinction between the motion planning of a vehicle and a mobile robot is that the vehicle is in a complex road network, and its route must also consider traffic regulations and road layout constraints, and in the process of network work, all routing algorithms must respond to network changes by refreshing the previous route. Common techniques include 1) graph search-based planners, 2) sampling-based planners, 3) interpolating curve-based planners, and 4) numerical optimization. The majority of current solutions consist of implementing route planning on a single server or dividing the road network into multiple processes using a parallel method. Nonetheless, deploying planning decisions directly in the cloud leads in communication latency, and at the same time, new collision trajectories are continuously generated due to multiple dynamic obstacles, which drastically reduces the decision window for route planning.
For the purpose of solving this problem, the work in\cite{Wilbur2020TimedependentDR} deploys routing at the network edge of RSUs and assumes that access to external cloud servers is intermittent while compressing the search space as well as limiting the communication frequency of fog nodes, effectively reducing latency and memory requirements.
 In \cite{Zeng2021MultiTaskFL}, the traffic data collected by each station is divided into distinct clusters, the road network is modeled as a time-dependent graph, and the enhanced A* algorithm determines the optimal route with the quickest travel time.

Furthermore, the utilization of UAVs as aerial anchors presents an opportunity to address the challenges associated with expensive and unstable GPS systems\cite{Liu2020PathPF}. Shahbazi \emph{et al.}\cite{Shahbazi2022FederatedRL} delve into this area by exploring local minimization error paths in various parameter environments. They employed FL to aggregate models and conducted tests in a fourth parameter environment, resulting in the identification of a model with the lowest localization error rate.
Regarding UAV route planning, it encompasses several considerations. Firstly, the presence of unforeseen flight obstacles, including enemy UAVs in military applications, must be taken into account. Currently, RL-based UAV route planning schemes have gained popularity. For instance, Khalil \emph{et al.}\cite{Khalil2022FEDUPFD} devise a route planning in hostile environments featuring dynamic and static defense systems through federated DRL. The inherent distributed learning structure of FL significantly enhances the performance of swarm-based models.
In addition, due to their limited power storage, UAVs often face resource constraints, necessitating optimal route planning achieved through joint power distribution and UAV scheduling methods. We will further explore this topic in the second subsection of Section \ref{4}.

The challenges in the route planning scenario can be summarized as the need to consider control constraints while perceiving the environment, with the optimization objective of minimizing the path or travel duration. In the current FL-enabled ITS architecture, utilizing the basic algorithms mentioned earlier to address route planning problems through iterative solutions can result in high computational costs. Additionally, as the number of vehicles in the system increases, it can lead to potential deadlocks.
Although there have been recent efforts to address dynamic route planning problems using reinforcement learning, which involves searching and trial-and-error methods to obtain suboptimal solutions, it also requires significant memory and bandwidth resources.

\begin{table*}[!t]
\begin{center}   
\caption{Application scenarios of FL in ITSs.} 
\renewcommand\arraystretch{1.2}
\begin{tabular}{|m{1.5cm}<{\centering}|m{3cm}<{\centering}|c|p{3cm}|p{4cm}|p{2.2cm}|c|}   
\hline     
\textbf{Scenario} & \textbf{Task} & \textbf{Ref}&\textbf{Algorithm} &\textbf{Limitations} & \textbf{Dataset} & \textbf{Year} \\  
\hline    
\multirow{5}{*}{\shortstack{Object\\Recognition}} & \multirow{3}{*}{Obstacle Detection} & \cite{Jallepalli2021FederatedLF} &YOLO V3 &The model aggregation mechanism is still simple. & KITTI &2021\\ 
\cline{3-7}    && \cite{Zhou2021TwoLayerFL}&CNN &Comparative experiments with single-layer FL are missing.& Private dataset & 2021 \\
\cline{3-7}    && \cite{Fang2021ResearchOU}&YOLO V3 &Impact of network structure not studied.& BelgiumTSC & 2021 \\
\cline{2-7}     &License Plate Recognition&\cite{Kong2021AFL}&YOLO, Mask-RCNN&The scenario of architecture deployment is relatively simple.&Private dataset&2021\\
\cline{2-7}     &Traffic Sign Recognition&\cite{Xie2022EfficientFL}& Spike Neural Network (SNN)&The security of the training algorithm is not taken into account.&BelgiumTSC&2022\\
\hline   
\multirow{11}{*}{\shortstack{Traffic Status\\Identification}} & \multirow{2}{*}{Vehicle Positioning} &\cite{Xu2020AnIT}& Single Shot Multibox Detector & The experiment is limited to static remote sensing images.& Remote sensing images from Google Earth & 2020 \\
\cline{3-7}   && \cite{Kong2022FedVCPAF} &DNN &Vehicle status information is not fully utilized to improve accuracy.& Didi Chuxing GAIA & 2022 \\
\cline{2-7}      &\multirow{6}{*}{Travel Mode Identification}& \cite{Torino2020FederatedLF}&LSTM & The simulation of LSTM is flawed. &Private dataset & 2020 \\
\cline{3-7}    && \cite{Gao2021GlowIT}&TCN (causal convolutions and the dilated
convolutions) & Ignored the label upload permissions.&Real-world crowdsourced dataset from DiDi & 2021 \\ 
\cline{3-7}   &&  \cite{Koetsier2021FederatedCD}&One-Class Support Vector Machine (OCSVM) &   Detection of a single anomaly class is missing. &INTERACTION Datase & 2021 \\
\cline{3-7}    && \cite{Nakanoya2021PersonalizedFL}&LSTM &Lack of communication overhead comparison.&  Private dataset & 2021 \\
\cline{3-7}    && \cite{Zhu2022SemiSupervisedFL}&CNN, GRU, ReLU &The server may not have data permissions.&  GPS data from the GeoLife project & 2022  \\
\cline{3-7}    && \cite{Zhang2021DistributedDM}&Knowledge Distillation &Assumptions on high accuracy of ego-vehicle localization.&  Public dataset & 2021 \\
\cline{2-7}     &\multirow{3}{*}{Traffic Flow Prediction}&\cite{Xia2022ShortTermTF}& GCN &Data heterogeneity is missing.& PeMS04, PeMS08 & 2022 \\
\cline{3-7}    &&\cite{Zhang2021FASTGNNAT}&GNN &Communication overhead is not taken into account.&  PeMSD7 & 2021 \\
\cline{3-7}  && \cite{Liu2020PrivacyPreservingTF}&GRU &Only security and privacy are concerned.&  PeMS & 2020 \\
\hline   
\multirow{3}{*}{\shortstack{Traffic \\ Management}} & Parking Management&\cite{Huang2021FedParkingAF}& LSTM & Only applicable to multi-leader game scenarios. &Birmingham parking dataset & 2021 \\
\cline{2-7}    &\multirow{2}{*}{Traffic Signal Control}& \cite{Wang2020AdaptiveTS}&Actor and Critic neural
networks &Communication latency is not taken into account.& Private dataset & 2020  \\
\cline{3-7}   &&  \cite{Ye2021FedLightFR}&Actor and Critic neural networks &The actual deployment is not considered.& Cityflow & 2021  \\
\hline   
\multirow{5}{*}{\shortstack{Service \\Providing}} & \multirow{6}{*}{Charging Service} &\cite{Wu2022ElectricVC}& FRF-CNN & Insufficient comparative experiments.& Public dataset & 2022 \\
\cline{3-7}   && \cite{Pokhrel2022DataPO}& RL &Data heterogeneity is not considered.& Private dataset & 2022 \\
\cline{3-7}   && \cite{Saputra2019EnergyDP}& DNN &Convergence speed results is not given.& Public dataset & 2019 \\
\cline{3-7}   && \cite{Wang2021ChargingSR}&  Encrypted entity alignment method &Communication overhead is not taken into account.& Public dataset & 2021 \\
\cline{3-7}   && \cite{Saputra2020FederatedLM}& Clustering-based DFEL &Lack of privacy protection mechanisms.&  Public dataset& 2020 \\
\cline{3-7}  && \cite{Thorgeirsson2021ProbabilisticPO}&Linear regression (LR) and neural network (NN) &Numerical simulations for privacy is lacked.&  TRDB & 2021 \\
\cline{3-7}  && \cite{Zou2023WhenHF}&HFL-LSTM, multi-agent double deep Q-learning & Analysis of performance loss from FL is lacked.&  Generated dataset & 2021 \\
\cline{2-7}   &\multirow{2}{*}{Route Planning}&  \cite{fWilbur2020TimedependentDR}& Partition Network &The handling of node failures is not considered.& OpenStreetMap & 2020  \\
\cline{3-7}   && \cite{Zeng2021MultiTaskFL}& A* &Traffic incident reporting incurs additional communication overhead.& PeMS & 2021 \\
\hline   
\end{tabular}   
\end{center}  
\label{table3}
\end{table*}

\section{Challenges in ITS based FL}\label{4}
In this section, we organize the application architecture of FL from the perspective of addressing the existing issues of ITS. We combine the challenges in Section \ref{1} into three aspects: high-speed mobility, limited storage and computing capability, and high information sensitivity. It will involve some commonly employed architectures in traditional FL, as well as, if necessary, adjustments for special scenarios. 
Notably, FL-enabled ITS alleviates the problems in traditional ITS scenarios to a certain extent, notably in terms of privacy protection and computational overhead reduction. Nonetheless, some problems that existed in FL are also present in ITS in the highly dynamic scenario, including higher communication overhead and latency. With the aid of additional mechanisms such as blockchain, UAVs, edge caching, etc., the existing problems can be greatly mitigated or even solved. These initiatives will facilitate the integration of FL with ITS.

\subsection{High-speed Mobility}
\subsubsection{Non-uniform Samples and Equipment}
As mentioned in Section \ref{1}, due to the highly dynamic topology of IoV, data is non-uniform and some vehicles are left behind. Both of the aforementioned factors inevitably result in non-uniform data and mobile node distribution (straggling vehicles are considered to drop out of the training process). Typically, to achieve low latency metrics for QoS, FL can enhance the tolerance for these issues by dynamically selecting participants and modifying the aggregation architecture\cite{Chen2020SemiFederatedL}. This idea is also applicable to the ITS scenario, but needs to be modified and adapted accordingly.

\textit{\textbf{Participant Selection:}}
Participant selection in traditional FL usually utilizes model quality, and model similarity, as a reference index for selection, and assigns lower aggregation weights to local models with poor performance or directly eliminates them, consequently decreasing the negative effect of poor samples\cite{Deng2021BlockchainAF}. Referring to this method, in ITS, numerous works replace the reference indicators with vehicle speed, sensor picture quality, and communication link quality to do multi-hop aggregation.

In\cite{Wang2021ContentbasedVS}, the authors take image classification as the scene, comprehensively consider loss function decay, wireless resources, computing resources, and energy as selection indicators for the first time, take learning efficiency as the final optimization goal, obtain vehicle selection and resource allocation schemes based on data content.
The work in\cite{Bao2021EdgeCJ} goes a step further, utilizing a fuzzy logic algorithm based on stability factor (vehicles with a relatively low speed can always be selected), topology factor, and connection factor (to ensure communication quality), considering vehicle speed, vehicle distribution and wireless link connections between vehicles, to select the appropriate edge vehicles for RSUs communication.
The measurement method adopted by \cite{Zhao2022ParticipantSF} in the selection mechanism is based on the contribution of the vehicles to the weight change of the model, i.e., the greater the weight change, the greater the contribution. Nonetheless, invariably, an increase in the amount of data can result in a greater contribution; however, this is accompanied by an increase in Non-IID, which impacts the training speed of each round. Accordingly, a correlation factor is added to the algorithm to balance this effect, and for the dynamic environment of IoV, the dimension of selection frequency is considered. Thus, based on the user's execution time and outlier detection, changes in the environment are identified.
Similarly, in\cite{Li2020SecureFL}, each vehicle participating in the training is required to confirm that its local update is consistent with the global update trend, and the resulting similarity is used to select vehicles.

In addition to the aforementioned techniques for extracting reference indicators based on the characteristics of the vehicle's environment, various conditions can also be combined into joint optimization problems, and subsequently the optimal solution can be obtained through a well-designed algorithm.
In \cite{Ye2020FederatedLI}, the authors employ the geometric relationship between the target and the camera to evaluate the image quality, in combination with the computing power of the vehicles, the model selection process is expressed as a two-dimensional contract theory, as well as the constraints are simplified by relaxation, and accordingly solved by a greedy algorithm.
In addition to vehicle selection, the work of\cite{Xiao2022VehicleSA} also takes into account resource scheduling, they treat resource allocation and vehicle selection as a joint optimization problem. First, with latency, transmission energy consumption, vehicle mobility (speed and position), and image quality as constraints, given the size of the vehicle selection set, the local training accuracy, on-board CPU frequency, as well as transmission power are optimized. Subsequently, the designed greedy algorithm adds vehicles iteratively to minimize energy consumption and latency (system cost).

Nonetheless, the above works ignore the issue of obtaining permissions for the cloud server's various indicators, that is, indicators including vehicle location, which users do not want to disclose to the server. In this regard, the research in\cite{Saputra2021DynamicFL} provides a feasible solution. The authors adopt an architecture for dynamic FL that allows a vehicular service provider (VSP) to access the location information of each smart vehicle (SV) securely. Moreover, to reduce privacy risks, this information is only used to assist the VSP in selecting the best SVs for each epoch of training. In other words, GPS information will be anonymized or obscured. Each collected SV determines the payment contract to the VSP based on its collected Quality-of-Information (QoI) and location significance. Moreover, to make the SV obtain the optimal payment contract, the author develops an optimization method on the basis of the MPOA protocol to deal with the common constraints of SV and VSP and the problem of insufficient payment budget.
This method generates additional information exchange between the server and clients, despite the fact that it prevents the server from directly obtaining private information. Therefore in\cite{Yang2022ClientSF}, the authors propose to hold a small validation dataset on the server side to measure the performance of selected client-updated models. And the dataset has ground-truth labels, and yet is insufficient to train a global model and does not incur excessive computational overhead.

In terms of theoretical analysis, the work in\cite{Zeng2021FederatedLO} presents a rigorous convergence analysis of its FL dynamic selection algorithm. Moreover, they analyze the effect of distinct participation in the FL process and distinct data quality on the convergence of the proposed algorithm.

\textit{\textbf{Aggregate Structure:}}
Some works redesign the aggregation architecture for FL in consideration of vehicle mobility. 
To adapt to the mobility characteristics of the vehicles, the work in\cite{Yu2021MobilityAwarePE} proposes a Mobility-aware Proactive edge Caching scheme on the basis of FL (MPCF), which selects a vehicle with sufficient computing power as well as storage capacity as a central server, termed server vehicle, rather than a static RSU, whereas nearby vehicles with the same driving mode are in connection to this server vehicle for training.
In\cite{Chen2021SemiasynchronousHF}, the authors refer to the standard FL approach to asynchronous problems, combining synchronous and asynchronous updates. And the architecture is divided into an edge layer and a cloud layer. Besides, the edge nodes aggregate a homogeneous local model, and the cloud layer aggregates a part of heterogeneous models. When the training of the slowest node is complete, the selected node updates the global model.

In terms of FL model updates, the work in\cite{Chen2021FederatedLW} studies the problem that each vehicle has distinct numbers of local images and the limitation of computing resources and communication resources. They propose an ``adaptive step/epoch" update method, in which each Non-IID participant's number of training steps for each epoch should be adaptively adjusted, to reduce training time while combining with model quantization to further reduce communication costs.

Some proposed strategies have not yet been implemented in ITS, including sharing a part of the original data between clients to improve the learning ability of Non-IID data\cite{Zhao2018FederatedLW}. Nonetheless, due to its reliance on data sharing, it can lead to serious privacy breaches. In\cite{Sahu2020FederatedOI}, it is suggested to regularize the local loss function using the squared distance between the local model and the global model, but the optimization efficiency is low.

Moreover, it should be noted that the method of using selection or client-side clustering is essentially an asynchronous update mechanism. This method is widely utilized in standard FL, but experiments indicate that some methods directly implemented in ITS scenarios will decrease performance\cite{Li2022FEELFE}. In addition, in dealing with the problem of uneven user resources, it is likewise possible to compensate for the performance loss of the stragglers or low-performance models via coding calculation, i.e., embedding calculation redundancy\cite{Prakash2020HierarchicalCG}.

\subsubsection{Unstable Transmission}
Another problem with the high dynamics of vehicles is that this mobility brings constant changes in network density, including areas where traffic congestion occurs with higher network density, and relatively remote areas have lower vehicle density. This density variation and distinct vehicle distribution lead to unstable information transmission, resulting in high communication latency.
 Aiming to achieve the end-to-end latency goal, it is essential to ensure low queue latency. Therefore, a significant amount of effort is devoted to the development of efficient wireless resource management schemes, including rate maximization, improving energy efficiency, vehicle network clustering, and power control algorithms.

It should be pointed out that the latency in ITS is predominantly caused by uneven computing capability and limited bandwidth. FL takes advantage of the computing power of edge nodes in a distributed manner, which effectively reduces the cloud server's computing power overhead, and yet the computing capability of edge nodes will be converted into a bottleneck; at the same time, since the vehicles only upload the model parameters, the transmission of the original data is avoided and the communication overhead is alleviated. Nonetheless, due to frequent communication between vehicles and RSUs, RSUs and the server, high communication costs continue to be incurred. In the sequel, we only discuss how to alleviate communication latency. And the computing latency is essentially caused by the limited computing capability, which will be discussed in the next section.

\textit{\textbf{Communication Protocol:}}
The two predominant current vehicle network distribution schemes are DSRC and cellular-based IoV\cite{Seo2016LTEEF},\cite{Bazzi2017OnTP}. This Vehicle-to-Everything (V2X) communication mode allows each vehicle to communicate with distinct entities, including RSUs, cloud servers, and other vehicles. Nonetheless, both schemes have shortcomings, DSRC with limited coverage, low transmission rates, and no mature safeguard mechanism for network security. Besides, cellular-based IoV cannot support distributed communication due to the fact that it can cause network congestion at high densities. Emergence of heterogeneous networks combining two communication protocols\cite{Zheng2015HeterogeneousVN}.

In \cite{Posner2021FederatedLI}, the authors equip each vehicle with DSRC and millimeter wave communication protocols in the designed IoV. The former is used for frequent communication of message control and small data, whereas the latter is utilized mass communication. In comparison with the communication protocol without distinction. It significantly enhances the system's scalability and stability as well as reduces communication latency 
In addition to DSRC and cellular networks, WiFi and other networks can likewise be incorporated into heterogeneous networks.
In \cite{Pokhrel2020ImprovingTP}, the authors design the transport protocol and network settings to capture highly dynamic on-Vehicle WiFi Access Points (oV-APs) data streams as well as link switching and retransmission due to packet loss or latency to precisely estimate packet loss rate as well as latency, and FL as either a collaborative management mechanism.

In terms of resource allocation, the work in\cite{Li2021FederatedLearningEmpoweredCD} proposes an information-sharing model on the basis of FL network routing optimization. Initially, the storage cost of each MEC is estimated, the energy consumption of MECs is calculated under the condition that the storage and latency constraints are satisfied, and eventually, the task scheduling is fed back to the target vehicles in order to decrease the system's total transmission latency.

Furthermore, the integration of air links and wireless powered communication (WPC) offers the potential to establish continuous communication for ground IoV. Zhou \emph{et al.}\cite{Zhou2018ComputationRM} propose the deployment of UAVs at near-ground locations to provide edge relay services for ground vehicles and user devices. They optimize the trajectory of the UAVs to enhance computational performance.
Moreover, Pham \emph{et al.}\cite{Pham2022EnergyEfficientFL} address the energy limitations of FL by leveraging UAVs. They consider UAV placement, power control, FL model accuracy, and bandwidth allocation as constraints to tackle the energy minimization problem in FL systems. By employing a path-tracing procedure to solve the convex approximation, they achieve convergence and optimize the system's energy efficiency.

\textit{\textbf{Link Structure:}}
Except for the optimization of communication protocols, there is also some works being done to redesign the network link structure.

Regarding that the single-hop cluster structure limits the system's coverage and stability, numerous clustering algorithms based on passive multi-hop have been developed in recent years\cite{Feteiha2014BestRelaySF},\cite{Zhang2019NewMC}. Using the network topology of ITS, the work of \cite{Li2022FEELFE} designs an end-to-end FL framework, as well as inter-cluster and inner-cluster learning algorithms, which accurately reduce redundant communication overhead. The above end-to-end approach, which can direct vehicles directly from raw data without costly labels, is widely used in ADS at present. In \cite{Pokhrel2020FederatedLW}, an autonomous blockchain-based FL design is proposed for privacy-conscious and efficient vehicular communication networks, and system-level performance is achieved by tuning parameters including block size, block arrival rate, and transmission limit, and rigorously analyzed and quantified end-to-end communication and consensus latency.
Hence, to collect more training models in vehicle cooperation perception, a blockchain-assisted propagation method based on FL is proposed in\cite{Ayaz2022ABB}. In smart contracts, the Proof-of-FL consensus competition transforms vehicles into miners. In comparison with smart contracts without security checks and other relay selection methods, the message propagation rate is dramatically improved.

As a distributed computing paradigm, IoT-Fog aims to place computational and storage resources at the edge of the network for low-latency and high-reliability task execution. First, one of the common communication types is the Proxy-Broker approach \cite{AlMasarweh2022FogCC}, in which the proxy and broker work in tandem to handle communication and information exchange. The proxy is employed to communicate with end devices, collecting and forwarding data to the broker, while the broker is responsible for processing, managing and routing the data. This communication type provides a flexible middle layer and improves the efficiency of data processing. Another type is the Publish-Subscribe approach\cite{Hernandez2022HandlingPA}, in which the publisher publishes data on multiple topics for subscribers to choose from. This communication type supports asynchronous communications, which is well suited for information dissemination and event-driven scenarios such as point-of-interest broadcasting in ITS. Later,\cite{Gao2022FogChainAB} investigates a P2P approach in localized networks (e.g., similar to the same cluster in FL) with low privacy requirements, wherein devices can communicate directly with each other.

Additionally, Software-Defined Networking (SDN), as a network architecture with programmability and centralized control system, consists of application, control, and data layers and is used to enable network device collaboration. Distributed deployment of fog servers with logical uniformity can be accomplished by using SDN enabled multi-hop IoT-Fog networks.
Akbar \emph{et al.}\cite{Akbar2020SDNEnabledAA} propose a combination strategy of machine learning and multi-objective optimization to search for optimal routes. It proves that SDN controllers can make adaptive decisions in real-time to select the best path from the Pareto-optimal set. However, in ITS, it is vital to consider network dynamic characteristics. Ibrar \emph{et al.}\cite{Ibrar2023ReliabilityAwareFD} utilize the Reliability-Aware Flow Distribution Algorithm (RAFDA) and two heuristic algorithms (RRAHA-1 and RRAHA-2) to allocate flows over links based on their reliability level.

When implemented in ITS, SDN supports a wide range of communication technologies, such as Bluetooth and ZigBee. Specifically, Bluetooth technology enables V2X communications that serve personal devices at a transmission rate of 3Mbps; ZigBee facilitates vehicle connectivity with sensors at a rate of 250Kbps\cite{Hakimi2021ASO}. 
At present, researchers advocate the integration of FL and SDN to improve FL's scalability \cite{Ma2022ApplyingFL}. Within this framework, SDN encompasses multiple local sub-controllers responsible for gathering network information such as traffic density and service type. A higher layer, called the master controller, plays the role of a central server in the ITS domain. It leverages standardized interfaces to aggregate local models and optimizes edge-to-edge routing paths for each service flow based on global network information. This integration enables not only region-wide communication but also customized routing configurations for diverse service flows to meet specific QoS requirements.

\subsection{Limited Storage and Computing Capability}
In comparison with traditional MEC scenarios, because vehicles in IoV have relatively huge energy reserves, the problem of limited transmission power is generally not considered, and yet there will be a large amount of data collection and storage in real-time. Additionally, although FL makes full use of the computing power of edge nodes through a distributed architecture, the storage and computing power limitations of the vehicles themselves must still be taken into account\cite{Chu2021FederatedLO}. And the existing solutions are usually optimization of resource scheduling or architecture design, including using Unmanned Aerial Vehicles (UAVs) as auxiliary mechanisms\cite{Masood2021ContentCI}, or proxy storage through third-party MECs and optimizing edge caching strategies.

\textit{\textbf{Auxiliary Mechanism:}}
In the MEC scenarios, the use of UAVs as an auxiliary mechanism is relatively mature, and a UAV-MEC system is quite suitable for dealing with high-density emergencies. Nonetheless, existing research often ignores distributed deployment and communication latency, and privacy issues.
The work in\cite{Nie2021SemiDistributedRM} proposes a multi-agent RL scheme under FL, which enables mobile devices to make offloading decisions based on the system's local condition, effectively saving power consumption. According to the authors, the algorithm is better suited for simulated environments and IoV scenarios.
In\cite{Aloqaily2021EnergyAwareBA}, the authors design a scheme integrating UAVs and SVs, ensuring the uninterrupted power supply of UAVs by considering the interaction of node power and mobility constraints to support the uninterrupted service of end devices.

However, the aforementioned research works fail to account for the impact of transmission distance on the delay and energy consumption of FL. Zeng \emph{et al.}\cite{Zeng2020FederatedLI} address this concern by optimizing the convergence rate of FL through power allocation scheduling. They also consider the energy consumption and delay of the swarm control system.
For UAV swarms operating within the FL architecture, a significant challenge arises from spectrum scarcity. To tackle this issue, Sabuj \emph{et al.}\cite{Sabuj2022APF} propose a UAV cognitive radio network (CRN) that utilizes local FL in the edge network to enhance spectral efficiency. Similarly, Wasilewska \emph{et al.}\cite{Wasilewska2021FederatedLF} aim to maximize spectral efficiency by exploring the relationship between computational and communication resources in FL-based CRNs.
Additionally, to enhance the energy efficiency of FL, Shen \emph{et al.}\cite{Shen2023JointTA} investigate the determination of the local convergence threshold and optimization of resource allocation to minimize the system's energy consumption in UAV swarms.

In addition to UAVs auxiliary resource optimization, edge servers can also be utilized as auxiliary agents. For instance, vehicles with limited computing resources can offload tasks to edge servers in time to decrease the computational load.
Even though ITS edge nodes can be deployed by third parties, there will be storage trust issues due to security risks\cite{Wu2018BigDA}. Therefore, blockchain is used as an auxiliary authentication mechanism in many works.
The work in\cite{Chai2021AHB} adopts a layered blockchain framework and Proof-of-Knowledge (PoK) lightweight consensus mechanism, and utilizes layered FL to introduce an intermediate layer to extract data correlation features, which reduces the amount of computation significantly.
In addition, Wang \emph{et al.}\cite{Wang2023DataIP} introduce DT into the ITS architecture and construct a twin city in the virtual space corresponding to the physical city to achieve intelligent management and decision making of the transportation system. They propose a conceptual model in environment-aware scenario.

\textit{\textbf{Edge Cache Optimization:}}
Optimization of edge caching is another popular research topic in ITS. Although the relatively close distance of RSUs can effectively reduce the connection latency, due to their limited computing resources, it is essential to optimize resource management by selectively offloading data. Passive and active caching solutions are available for the edge. Passive caching selects cached content by observing the pattern of user requests. This method can only be cached after the request, making it vulnerable to obsolescence in environments with a high level of dynamic IoV. To achieve a sufficiently high cache hit rate, a large amount of backup storage must be performed, which increases the storage pressure. And the active caching strategy on the basis of content prediction is very suitable to solve this issue. In the training of the active caching model, the work in\cite{Yu2021MobilityAwarePE} allocates the FL aggregation weight in accordance with the vehicle storage resources, which optimizes the resource utilization.

Edge caches, like vehicles, must not only ensure a high cache hit rate, but also perform precise computing offloading to further alleviate computing pressure. Whereas using the FL framework can learn to offload the information from the decision-making stage, which incurs a greater cost in terms of latency, despite the fact that it can help make better decisions.
In\cite{Shinde2022OnTD}, three RSU-based clustering methods and vehicle-based distributed methods are designed to model the unloading process by delay and energy consumption, and the delay and energy minimization problem is modeled as an optimization problem under nonlinear constraints. The ultimate optimization objective is to select an appropriate number of FL iterations for each vehicle based on the latency requirement in order to minimize energy consumption.

Additionally, not only the information is cached in RSUs, but also a small part of the works have studied how to cache the information in vehicles. Currently, the stability of dynamic storage must be thoroughly considered.
The work in\cite{Hu2019InVehicleC} proposes a dynamic distributed on-board storage system, to prevent the problem of data loss caused by vehicles leaving the management area. They set up data transmission areas at the entrance and exit of each area, so that the stored data can be transmitted from of the departing vehicle to the arriving vehicle via a one-hop link in a timely manner. Furthermore, the authors maintain the integrity of the storage by introducing structural redundancy of erasure codes.

Note that while edge arithmetic offloading reduces communication latency and the computational burden on vehicles, it also introduces greater network complexity and computational redundancy. To address these challenges, the combination of SDN and MEC emerges as a promising approach. Local SDNs serve as control systems responsible for managing task migration between edge servers\cite{Zhuang2020SDNNFVEmpoweredFI}. These local controllers gather vehicle state information through cellular networks or DSRC, enabling more effective task offloading policies. Recently, distributed SDN controllers have been employed as agents, as seen in the work of Tam \emph{et al.}\cite{Tam2022OptimizedMT}, to mitigate update degradation by executing tasks on nodes with sufficient computational capacity. They also recommend suitable edge FL servers for model aggregation and averaging. Chen \emph{et al.}\cite{Chen2019AJL} investigate various parameters influencing local FL model transmission, optimizing these parameters based on wireless network characteristics. Balasubramanian \emph{et al.}\cite{Balasubramanian2021IntelligentRM} additionally considers cache location and content privacy. 
Local device data is used to train the model, and local updates are then uploaded to the central controller. The FL module determines content placement based on QoS, and the SDN controller adjusts the routing path accordingly. Overall, SDN-assisted FL can be deployed in dynamic network topologies, facilitating trusted resource management.

It is important to note that FL primarily serves as a privacy-preserving mechanism in edge caching. Additionally, FL eliminates the need for centralized collection of training data, thereby enhancing caching efficiency\cite{Cheng2021APD}.
Ultimately, there are a part of the works using deep Reinforcement Learning (DRL) for optimizing resource allocation, which is effective in improving energy efficiency, and yet for lightweight edge devices, especially edge vehicles, in most cases, it is insufficient to support the deployment of the RL framework independently\cite{Chen2019OptimizedCO}.


\subsection{High Information Sensitivity}
Vehicles are currently more susceptible to tracking attacks than mobile phones, making privacy protection mechanisms in ITS a particularly pressing issue. In the first place, due to the requirements for vehicles safety, the vehicles communicate frequently with RSUs to ensure real-time service on the basis of positioning, which substantially increases the likelihood of being attacked. Second, the vehicle moves under traffic control and road constraints, and it is much simpler to attack the position of vehicles with logic than it is to predict the movement of the MEC as a whole.
Notable examples of such attacks include eavesdropping attacks and location attacks. To elaborate further, the Global Passive Adversary (GPA) exploits valid certificates to glean vehicle information, including user IDs, location information, RSU details, license plate readers, as well as comprehending road structure and traffic conditions. Although the GPA attack is primarily regarded as a passive attack, entailing the interception of private data without direct tampering, it is often executed on a global scale, with the GPA leveraging traffic management systems to achieve broad data coverage.
FL, on the other hand, as a distributed paradigm with third-party agents, protects the privacy of participating vehicles to a certain extent due to the fact that the original data is mostly kept local\cite{Wei2021UserLevelPF}. Nonetheless, some malicious attackers can nevertheless deduce the original data backward from the uploaded gradient information\cite{Xu2019DataSI}. Moreover, many works do not account for the existence of malicious vehicles or servers, including malicious vehicles will upload wrong model parameters to interfere with the training process and affect training accuracy. Consider a scenario involving road congestion, where an attacker manipulates the congestion signal by tampering with the data, thereby altering the navigation path. While some research suggests that pseudonyms can be employed to safeguard vehicle data privacy\cite{Hussain2014CooperationAwareVC}, this approach can still be compromised in the absence of adequate trust between cloud members. An attacker can infiltrate the system by posing as a member of the cloud, thereby necessitating the exploration of dynamic trust mechanisms as a potential area of investigation.

Furthermore, the traditional FL architecture proves ineffective against certain program attacks targeting channel allocation. Examples of such attacks include denial of service attacks, where dishonest vehicles impede all feasible communication methods by transmitting multiple messages. These attacks can be executed in a distributed manner, known as distributed denial of service attacks (DDoS). Additionally, jamming attacks disrupt broadcast communications through various techniques such as alarm injection, which consumes the available frequency range. Jammers impede network communications between vehicles within a designated transmission and reception range by interfering with the physical transmission and reception of wireless communications.
\subsubsection{Blockchain}
ITS inherently face security challenges related to availability, identity recognition, and confidentiality. Conventional identification systems rely on public-private key mechanisms, necessitating vehicle authentication with local certification centers. However, this periodic encryption and decryption process introduces additional network overhead. In contrast, blockchain, as a decentralized paradigm, offers a solution to mitigate attacks like single-point attacks. It operates similarly to entity-centric vehicle trust management, establishing a unified reputation system or making decisions based on neighboring user opinions\cite{Shi2021PoolingIN}. By integrating FL and blockchain, some privacy concerns can be addressed, and issues related to untrusted data storage, centralized trust, and tampering in existing systems can be resolved\cite{Du2023BlockchainAidedEC}. Recent research has proposed architectures that combine FL and blockchain to tackle these challenges in ITS.
The work in\cite{Lu2020BlockchainEA} proposes an FL architecture on the basis of a hybrid blockchain architecture consisting of a permissive blockchain and a locally directed acyclic graph, employing DRL for node selection, integrating the learned model into the blockchain and conducting two-phase verification.
Nonetheless, there are a large number of P2P communication methods in ITS, and direct blockchain coupling in this scenario will reduce efficiency\cite{Lu2020BlockchainEA}.
In\cite{Chen2021BDFLAB}, the authors provide a P2P networks structure that is completely independent of the blockchain and deploys a publicly verifiable secret sharing scheme to protect data without having to compromise model precision.
In UAV scenarios, there are security concerns when using traditional blockchain and FL architectures for data analysis due to the open nature of UAV communications. To address this, Zhu \emph{et al.}\cite{Zhu2022BlockchainEnabledFL} propose a blockchain-assisted MEC framework for FL data circulation. In this architecture, the global model is distributed on the blockchain and resides on UAVs. The UAVs upload model parameters to the MEC server through wireless networks, where the MEC server adds digital signatures to the model and broadcasts it. Other servers receive the local model, perform mining and block verification operations.

While blockchain-assisted FL can address single-point failures and communication overhead between vehicles and the main server through decentralization, the primary communication costs still concentrate on edge computing and parameter transmission delays\cite{Li2021BlockchainAD}. Current solutions focus on joint optimization considering various latency requirements and can improve communication efficiency through gradient compression techniques while enhancing data privacy protection.
It is important to note that the protection mechanisms provided by blockchain are based on anonymous identity authentication strategies, which are suitable for preserving identity privacy. However, they overlook the protection of model parameters and are still susceptible to threats such as data and model poisoning, as well as issues like transmission security and location privacy protection\cite{Wiedersheim2010PrivacyII}. Specifically, before submitting the aggregation, an attacker can manipulate local training and hyperparameters. Although adjusting the mining difficulty can mitigate poisoning risks, it also entails performance trade-offs.

\subsubsection{Encryption Mechanism}
A lot of work at this stage is based on the FL architecture in ITS to do cryptographic mechanism design.

\textit{\textbf{Differential Privacy:}}
To further protect model parameters in FL, some studies have added DP mechanisms. DP signifies that the vehicle independently executes a random perturbation algorithm and then sends the perturbed result to the aggregator, where the privacy budget parameter is used to calculate the privacy utility\cite{Wei2019FederatedLW}.
The research in\cite{Zhao2021LocalDP} proposes four new local differential privacy (LDP) mechanisms, PM-OPT, PM-SUB, Three-Outputs, and HM-TP. In comparison with the existing LDP mechanism, they have lower worst-case noise variance in the segmented privacy budget range. For the purpose of facilitating coding, they discretize the continuous output range. Thus it is more suitable for ITS scenarios.
In addition to introducing noise into the original data and parameters, the authors in\cite{Zhang2021FASTGNNAT} introduce a DP-based adjacency matrix in the proposed FL framework to protect topological information (which may likewise contain sensitive information, including the relationship between data providers).
Furthermore, the article combines blockchain and DP, using delegated practical byzantine fault tolerance (dBFT) model to update the validation mechanism to reach a consensus, in order to guarantee that only qualified models are aggregated. And the LDP mechanism prevents membership inference attacks by adding Gaussian noise to the model.
In theoretical analysis, the work in\cite{Li2020SecureFL} demonstrates model convergence under noise. 

Nonetheless, DP is not a perfect solution for FL to protect vehicle privacy\cite{Jayaraman2019EvaluatingDP}. In the first place, it is necessary to find an appropriate DP factor to adjust the degree of privacy protection. Furthermore, this way of adding noise is lossy, and ITS has an extremely low accuracy tolerance, notably in scenes involving vehicle operation. People are opposed to sacrificing driving safety for privacy protection.

\textit{\textbf{Other Encryption Mechanism:}}
In ITS, it is necessary to consider employing lightweight, low-loss encryption algorithms.
For example, Parekh \emph{et al.}\cite{Parekh2023GeFLGE} use the computational power of edge devices to fine-tune the local model and perform gradient encryption to save arithmetic overhead based on a traffic sign classification scenario.
In addition to designing the encryption algorithm itself, it is also possible to allocate weights in accordance with the privacy needs of the packet and to allocate encryption resources confirming to the execution time.

 The authors in\cite{Peng2021BFLPAA} choose distinct FL methods according to the distinct distribution characteristics of the data source, and adopts a lightweight encryption algorithm CPC to protect privacy, in comparison with other symmetric encryption algorithms, which eliminates the expense of computation.
In \cite{Li2022PrivacyPreservedFL}, the authors design encryption algorithms for two distinct MEC servers. Among them, for semi-honest vehicles, the use of an identity anonymity scheme to protect message privacy, through the use of a key of length 8, can achieve nearly 99$\%$ privacy protection in a single full authentication. For malicious vehicles, an identity traceability scheme is employed, as well as blockchain-based autonomous driving reputation incentives are being used to mitigate the negative effects of malicious vehicles by increasing vehicle contribution.
In \cite{Yamany2021OQFLAO}, the authors propose an optimized quantum-based FL framework for automatically tuning FL's hyperparameters, including local epoch, global epoch, as well as learning rate in the face of adversarial attacks, then find the optimal solution using the suggested optimization algorithm.

It should be noted that despite the fact that homomorphic encryption (HE) is a commonly used encryption scheme in traditional FL, it requires a tremendous amount of computing resources. Even linear homomorphic encryption (LHE) requires complicated modular exponential operations, particularly for large-scale networks like ITS, it is impractical to use FL on the basis of HE.

Moreover, in the previously mentioned IoT-Fog network, fog devices are usually placed near IoT devices in a distributed manner without security defenses. As a result, the fog layer is highly vulnerable to malicious attacks\cite{Javanmardi2023AnSP}. Traditional techniques for network security, such as intrusion detection systems, black-and-white-list mechanisms, and firewalls, are ineffective in the large-scale IoT-Fog paradigm due to distributed, dynamic, and heterogeneous FL-enabled ITS. These approaches do not provide security for borderless system IoT-Fog architecture, because nodes in the fog layer can be easily changed through local or remote connections\cite{Zaminkar2020AMB}. In contrast, SDN can provide programmable network and global view for efficient traffic management in FL-enabled ITS by deploying centralized SDN controllers to observe the nodes in cloud servers. Gao \emph{et al.}\cite{Gao2020ABI} incorporate blockchain and SDN into ITS to run the 5G fog computing paradigm in distributed scenarios. Fog computing can be used to mitigate handoffs between vehicles, and blockchain provides a decentralized reputation-scoring mechanism to reduce the risk of attacks, rather than being centralized on a single institution or server. In their proposed architectures, the SDN data plane selects channels in vehicles equipped with SDN-enabled in-vehicle devices, roadside units, and base stations. The distributed RSUs act as SDN controllers and perform the blockchain operations. ELMansy \emph{et al.}\cite{Elmansy2022MPTCPbasedSS} propose a lightweight mitigation system for fog computing in response to the Man-in-the-Middle (MITM) attack in fog computing networks. The system utilizes the available connectivity interfaces in the Edge Devices and Fog Open vSwitches to provide redundant paths for edge devices and Fog nodes. In light of SDN deployment for FL-enabled ITS, Qureshi \emph{et al.}\cite{Firouzi2022ADS} study a distributed SDN approach in a smart grid scenario. Each distributed controller is responsible for a specific domain, obtains a partial view of the network, and finally updates the global state. In FL-enabled ITS, SDN supports efficient segmentation of the network. The different components of the ITS infrastructure are divided into logical segments to prevent unauthorized access and tampering. In addition, the programmability of SDN enables dynamic enforcement of security policies based on real-time network conditions. Nevertheless, in order to achieve the security goals of SDN-based IoT-Fog networks, it is essential to take into account the uneven channel quality and resource constraints as well as the diverse network topologies of ITS\cite{Deb2022ACS}.

\subsubsection{Detection Mechanism}

In addition to using encryption algorithms, privacy is also protected by designing defense mechanisms including attack detection and data leakage detection. In \cite{Boualouache2022FederatedLS}, a FL-based architecture for detecting passive attackers who eavesdrop on vehicles' information is proposed. Besides, the author first simulates passive attackers through synthetic data and position-feature extraction method, and consequently uses a semi-supervised method to self-label data in FL vehicles to obtain precise detection results in a short amount of time. The authors in\cite{Lu2020FederatedLF} devise a two-stage mitigation scheme including data transformation (converting raw data into a data model) and collaborative data leakage detection (by a DP model distorting and randomly selecting updates to prevent the model's differences from being used to infer provider information).

\begin{table*} [!t]
\begin{center}   
\caption{Challenges in ITS based FL.} 
\renewcommand\arraystretch{1.2}
\begin{tabular}{|m{1.6cm}<{\centering}|m{1.4cm}<{\centering}|m{1.3cm}<{\centering}|c|p{5cm}|p{4cm}|c|}   
\hline     
 \textbf{Characteristic} & \textbf{Challenge} & \textbf{Method} &\textbf{Ref}& \textbf{Algorithm / Framework}& \textbf{Contribution / Performance} & \textbf{Year} \\  
\hline    
 \multirow{18}{*}{\shortstack{High-speed \\Mobility}} 
& \multirow{12}{*}{\shortstack{Unbalanced \\samples\\and\\ equipment}} 
& \multirow{9}{*}{\shortstack{Participant\\ Selection}} &\cite{Bao2021EdgeCJ}& Fuzzy logic algorithm &A trade-off between accuracy and communication overhead &2021\\ 
 \cline{4-7}     &&& \cite{Wang2021ContentbasedVS}&Genetic algorithm&More accurate, converge faster&2021\\
\cline{4-7}     &&&\cite{Zhao2022ParticipantSF}&Selection frequency & Model accuracy increases 20$\%$&  2022\\
\cline{4-7}     &&&  \cite{Li2020SecureFL}&&Convergence proof, less 
communication overhead & 2020\\
\cline{4-7}     &&& \cite{Ye2020FederatedLI}&Two-dimension contract theory, greedy algorithm&Higher utility at the central server  &2020 \\
\cline{4-7}     &&&  \cite{Xiao2022VehicleSA}&Greedy algorithm, subgradient projection method, adaptive harmony algorithm & A tradeoff between training time and
energy consumption&2022 \\
\cline{4-7}    &&& \cite{Saputra2021DynamicFL}& MPOA-based contract, learning contract iterative algorithm  &  Convergence proof, converge faster& 2021\\
\cline{4-7}     &&& \cite{Yang2022ClientSF}&Copeland score and multi-arm bandits framework&Converge faster &2022 \\
\cline{4-7}     &&&\cite{Zeng2021FederatedLO}&Dynamic federated proximal (DFP) algorithm & Convergence proof, converge 40$\%$ faster&2022 \\
\cline{3-7}    && \multirow{3}{*}{\shortstack{Aggregate\\ Structure}} & \cite{Yu2021MobilityAwarePE}&Mobility-aware Proactive edge Caching scheme &Cache hit rate improves about 2$\%$&2021\\  
\cline{4-7}     &&&\cite{Chen2021SemiasynchronousHF}&Alternating Direction Method of Multipliers
(ADMM)-Block Coordinate Update (BCU) algorithm &A tradeoff between accuracy and transmission latency & 2021\\
\cline{4-7}     &&&\cite{Chen2021FederatedLW}&``Adaptive steps/epoch"& 66$\%$ training faster and reducing 35$\%$ communication cost & 2021\\
\cline{2-7}     & \multirow{6}{*}{\shortstack{Unstable \\Transmission}} & \multirow{3}{*}{\shortstack{Communi-\\cation \\Protocol}}   &\cite{Posner2021FederatedLI}& Federated vehicular cloud (FVC)& Reduce average latency&2021\\ 
\cline{4-7}     &&& \cite{Pokhrel2020ImprovingTP}&TCP CUBIC over WiFi-based network&Theoretical analysis & 2020\\
\cline{4-7}     &&&\cite{Li2021FederatedLearningEmpoweredCD}&DQN &Lower latency &2021\\
\cline{3-7}     && \multirow{3}{*}{\shortstack{Link\\ Structure}}   & \cite{Li2022FEELFE}&End-to-End framework, Paillier-based communication protocol& Converge faster&2022\\  
\cline{4-7}     &&& \cite{Ayaz2022ABB}& Proof-of-FL (PoFL) consensus & Reduce 65$\%$ latency,  improve 8$\%$ message delivery rate &2022\\
\cline{4-7}     &&& \cite{Pokhrel2020FederatedLW}&oVML algorithm& Theoretical analysis, lower latency &2020\\
\hline   
\multicolumn{2}{|c|}{\multirow{7}{*}{\shortstack{Limited Storage \\and \\Computing Capability}} }
& \multirow{3}{*}{\shortstack{Auxiliary\\Mechanism}} &\cite{Nie2021SemiDistributedRM}& MARL approach&Low power consumption &2021\\ 
\cline{4-7}     \multicolumn{2}{|c|}{}&& \cite{Aloqaily2021EnergyAwareBA}&Cooperative
UAV-UGV process &Low power consumption &2021\\
\cline{4-7}     \multicolumn{2}{|c|}{}&&\cite{Chai2021AHB}&PoK consensus mechanism, alternating direction method of multipliers-based  &Model accuracy increases 20$\%$ &2021\\
\cline{3-7}    \multicolumn{2}{|c|}{}& \multirow{4}{*}{\shortstack{Edge Cache\\ Optimization}}   & \cite{Shinde2022OnTD}&Evolutionary Genetic Algorithm& Lower energy consumption&2022\\
\cline{4-7}     \multicolumn{2}{|c|}{}&&\cite{Hu2019InVehicleC}&  Data redundancy by erasure coding and data relay by V2V communications&Theoretical analysis, the system could serve up to 95$\%$ of the requests for contents  &2019\\
\cline{4-7}     \multicolumn{2}{|c|}{}&&\cite{Chen2019AJL}&  Enabling the implementation of FL algorithms over wireless networks &Framework can improve the identification accuracy by up to 1.4\%, 3.5\% and 4.1\%  &2019\\
\cline{4-7}     \multicolumn{2}{|c|}{}&&\cite{Balasubramanian2021IntelligentRM}&  SDN-controlled FL framework &Not only provides secure and trustworthy service delivery but also ensures seamless communication &2021\\
\hline   
\multicolumn{2}{|c|}{\multirow{11}{*}{\shortstack{High Information Sensitivity}} }
& \multirow{2}{*}{\shortstack{Blockchain}} & \cite{Lu2020BlockchainEA}& Local Directed Acyclic Graph (DAG), blockchain two-stage verification&More accurate and converge faster &2020\\ 
\cline{4-7}    \multicolumn{2}{|c|}{}&&\cite{Chen2021BDFLAB}&HydRand protocol, PVSS scheme & More accurate, converge faster &2021\\
\cline{3-7}      \multicolumn{2}{|c|}{}& \multirow{4}{*}{\shortstack{Differential \\Privacy}}   &\cite{Zhao2021LocalDP}& Four new LDP schemes& Protecting privacy while guaranteeing utility &2021  \\
\cline{4-7}      \multicolumn{2}{|c|}{}&&\cite{Zhang2021FASTGNNAT}&Add noise to the adjacency matrix &Satisfy DP &2021\\
\cline{4-7}     \multicolumn{2}{|c|}{}&& \cite{Qi2021PrivacypreservingBF}& delegated Practical Byzantine Fault Tolerance (dFTP)&Preventing data poisoning attacks &2021\\
\cline{4-7}     \multicolumn{2}{|c|}{}&&\cite{Li2020SecureFL}&Identifying relevant updates trained by vehicles & Convergence proof, 4.0x communication efficiency &2020\\
\cline{3-7}      \multicolumn{2}{|c|}{}& \multirow{3}{*}{\shortstack{Others}}  & \cite{Peng2021BFLPAA}&CPC encryption&Lower calculation cost&2021\\ 
\cline{4-7}     \multicolumn{2}{|c|}{}&& \cite{Li2022PrivacyPreservedFL} &Traceable and anonymous  identity-based scheme&Reducing 73.7$\%$ training loss, increaseing 5.55$\%$ accuracy &2022\\
\cline{4-7}     \multicolumn{2}{|c|}{}&&\cite{Yamany2021OQFLAO}&Optimized Quantum-based FL & Highest detection accuracy& 2021\\
\cline{3-7}     \multicolumn{2}{|c|}{}&\multirow{2}{*}{\shortstack{Detection \\Mechanism}} & \cite{Boualouache2022FederatedLS}&Detect tracking attacks based on the receiving beacons& 20 received beacons could achieve 95$\%$ accuracy& 2022\\
\cline{4-7}      \multicolumn{2}{|c|}{}&&\cite{Lu2020FederatedLF}&Intelligent data transformation and collaborative data 
leakage detection & Data leakage 
defending scheme is near-real-time&2020\\
\hline   
\end{tabular}   
\end{center}  
\label{table4}
\end{table*}

\section{Challenges and Future Research Directions}\label{5}
Despite researchers have made significant efforts on FL-based ITS solutions and achieved remarkable results, in addition to the numerous difficulties mentioned previously, there are some issues worth considering, and in this section, from a broader perspective, we extend our thinking and propose several future research directions for researchers to consider:
\begin{itemize}

\item \textbf{Lightweight Encryption Algorithms:} 
Numerous works on vehicle privacy protection, including identity anonymity and multiple encryption techniques, have been published. However, many encryption algorithms incur additional computational overhead, including key-based encryption technology, which needs to consume a lot of computing resources to manage keys. Few lightweight encryption algorithms are specifically designed for ITS, according to our survey. Meanwhile, there is a lack of performance testing and experimental verification in real scenarios, i.e., there are very few works to test their defense models with distinct attack types, including backdoor attacks, symbol flipping attacks, etc.  

\item \textbf{Privacy and Security Concerns of FL:}
Although the principle of ``no data out of local" mitigates privacy issues to some extent in FL, attackers can often deduce individual data pertaining to participants' vehicles using model parameters or gradient information. While encryption techniques such as DP can provide masking protection, they may compromise decision accuracy, which is often unacceptable in ITS\cite{Wei2021LowLatencyFL}. Furthermore, regarding security, although blockchain can mitigate single points of failure, attackers can manipulate the distribution of training data by strategically inserting crafted samples into the training set. Existing strategies, including participant authentication and other techniques, are all limited by the additional communication overhead and system management costs.
\item \textbf{Enhancing Incentives and Selection Mechanisms:}
In FL-enabled ITS, there is an inherent imbalance in the information shared, such as computing power and storage capacity. Therefore, it is imperative to establish suitable incentive schemes that accurately assess vehicle contributions. While there are currently numerous incentive mechanisms and vehicle selection methods, the lack of a standardized system results in divergence among incentive schemes employed in different systems or applications, thus blocking compatibility and reusability.
\item \textbf{Scarcity of Labeled Samples in FL:}
Although FL can maximize the use of local samples through model interoperability, the availability of labeled training samples is limited due to distinct vehicle sensor performance and variable environmental conditions. Data augmentation techniques can expand the sample pool, while potentially introducing noise or human intervention. Transfer learning can leverage data from different domains, but it must address inter-domain discrepancies and domain adaptation challenges to ensure optimal model performance within the target domain.
\item \textbf{Real-world Architecture Deployment:}
The combination of FL and RL is a hot topic in signal lamp control and route planning\cite{Haydari2022DeepRL}, and despite experiments have shown that this scheme can learn nonlinear optimal strategies, most existing models cannot be applied to real-world environments. Consequently, on the one hand, how to deploy in the actual scene should be taken into account, on the other hand, the performance of the vehicle simulator are further enhanced, and some random human intervention should be added to ensure generalization ability in the actual scene.   
\item \textbf{FL Application in ITS Generalized Scenarios:} 
We observe that almost no ITS scenarios exist in isolation, and each scenario can affect the others, including the goal recognition task can be used as a subset of the route planning task, and the constraints in the route planning will be distinct depending on the purpose of the task, including the shortest travel distance, the best resource allocation, the best traffic flow control, etc. We consider whether distinct sub-tasks can be promoted to a broader scene in the future. In this scenario, the same FL is used as the underlying service architecture.
\item \textbf{More Efficient Sensor Configuration:} 
The majority of tasks in ITS rely on a large number of sensors, and the data types and storage methods collected by these sensors are not the same. Besides, how to measure the contribution of distinct sensor data to the current task? Do we need to customize networking and storage for distinct data and tasks to maximize benefits? These are all worthy of consideration.
\item \textbf{Trade-off between Privacy and Performance:} 
IoV's security and privacy has always been one of the most important concerns. Additional resource allocation is required in some private computing and message verification links. Particularly crucial is determining how to allocate computing resources rationally and strike a reasonable balance between privacy protection and model precision.
\item \textbf{Edge Cache Authentication:} 
 In the event that a vehicle has limited computing capabilities, we note that a portion of the computing tasks can be offloaded to edge nodes. Static nodes like RSUs, on other hand, do not move with the vehicles, which can result in the leaking of information to newly entering vehicles. Additionally, some auxiliary nodes are provided by third parties and cannot be relied upon, i.e., the training results provided by RSUs are not completely trusted. In the future, we can consider designing a lightweight verification mechanism to ensure that edge processing results are verifiable.
 \item \textbf{Reduce Transmission Overhead:} 
 In the previous section, we mentioned that by optimizing the communication link, network robustness and communication latency can be enhanced. Additionally, compression-based methods are often used in traditional FL to reduce the amount of data and transmission overhead, including pruning and quantization. In model compression, the optimal model structure needs to be designed, and distinct communication standards can be blended to improve the overall network reliability based on FL.
\item \textbf{Deep Integration of ITS and FL:} 
Although FL has been widely used in ITS, there are few works dedicated to the design of special FL architectures for distinct data types and scene characteristics used for distinct tasks. We hope that future research will consider increasing the degree of coupling between FL and ITS, proposing some better paradigms, and focusing more on the design and implementation of the underlying architecture to achieve deep integration of sensors, network computing, and vehicle network integration.  
\item \textbf{Commercial Products for ITS:} 
Mature commercial platforms can be explored for deployment in ITS in the future. For example, OpenDaylight, an open source SDN controller platform, can provide ITS with rich network programming interfaces for supporting interoperability of multiple communication protocols and data formats. It can also be integrated with other security solutions to provide security management and protection. Similarly, Cisco Kinetic, an IoT management platform provided by Cisco, can be deployed in ITS to manage large numbers of devices.
\end{itemize}

\section{Conclusion}\label{6}
In this paper, we have investigated the integration of FL and ITS. Analyzing several problems of centralized ITS and the advantages of FL, we presented the reasons for using FL architecture in ITS. Subsequently, we conducted a comprehensive survey on the applications of FL in ITS and identify four distinct application scenarios. 
Then, from the viewpoint of ITS characteristics and in conjunction with related works, we elaborated on how the FL paradigm can address key challenges in ITS, including uneven data distribution, limited computing and storage resources, limited transmission, and data privacy and security.
We also discussed security mechanisms of the blockchain and UAV-aided FL resource scheduling. Finally, we highlighted open problems and proposed future research directions. 
It is worth noting that due to the page limit, several enabling technologies in ITS, such as integrated sensing, mobile fog computing, cluster routing, deserve further discussion.

\bibliographystyle{IEEEtran}
\bibliography{IEEEabrv,Ref}


\begin{IEEEbiography}[{\includegraphics[width=1in,height=1.25in,clip,keepaspectratio]{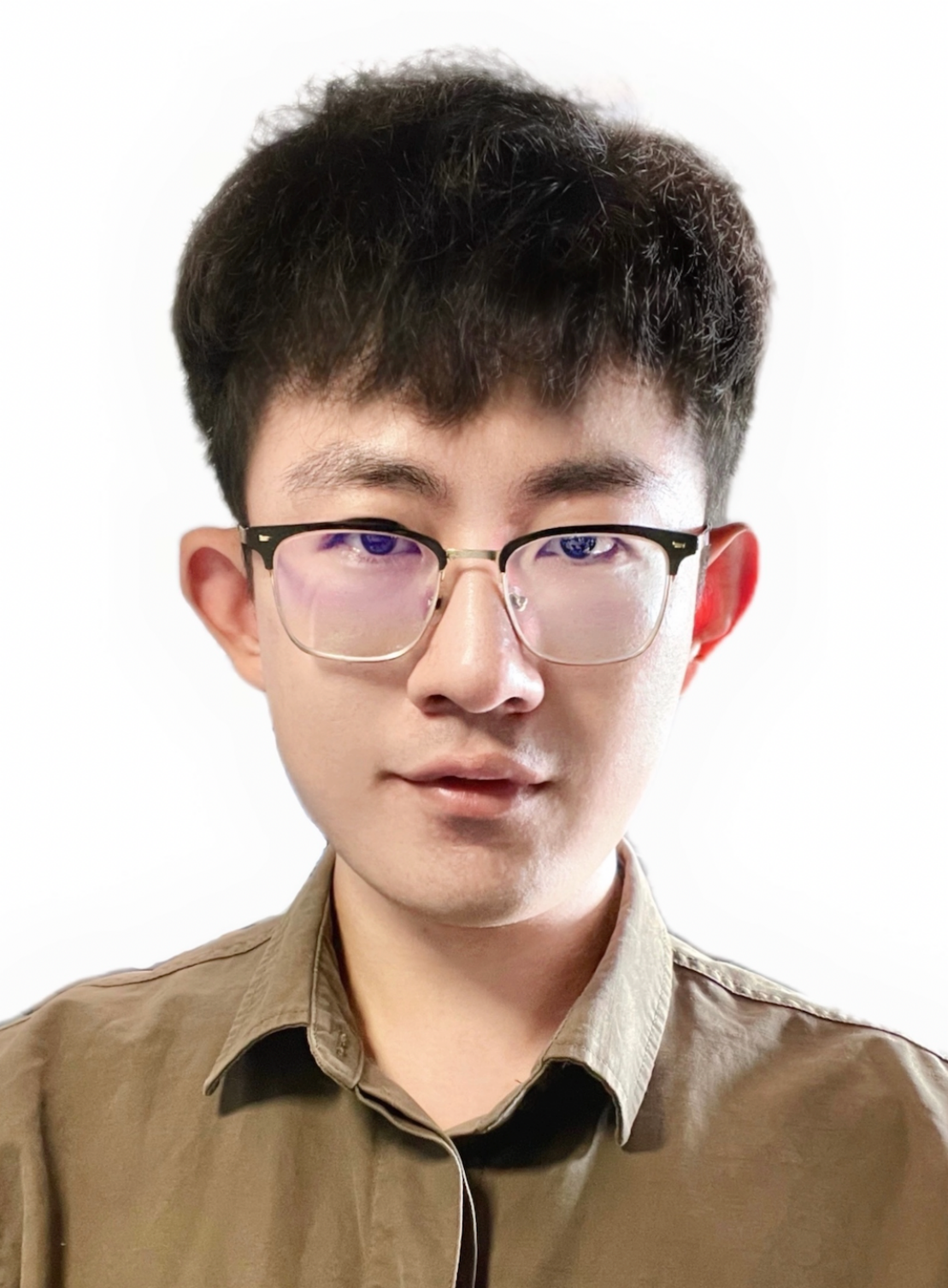}}]{Shiying Zhang}
 is currently pursuing the Ph.D. degree in communication and information system with the School of Electrical and Optical Engineering, Nanjing 210094, China. His research interests focus on federated learning, deep learning and mobile
edge computing, security and privacy.
\end{IEEEbiography}
\begin{IEEEbiography}[{\includegraphics[width=1in,height=1.25in,clip,keepaspectratio]{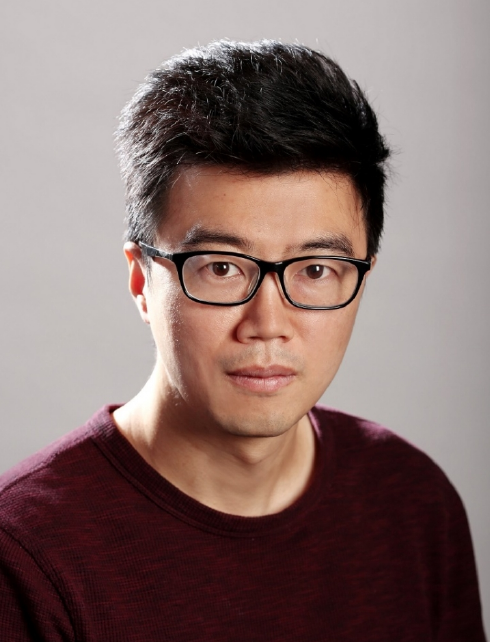}}]{Jun Li}
 (Senior Member, IEEE) received Ph. D degree in Electronic Engineering from Shanghai Jiao Tong University, Shanghai, P. R. China in 2009. From January 2009 to June 2009, he worked in the Department of Research and Innovation, Alcatel Lucent Shanghai Bell as a Research Scientist. From June 2009 to April 2012, he was a Postdoctoral Fellow at the School of Electrical Engineering and Telecommunications, the University of New South Wales, Australia. From April 2012 to June 2015, he was a Research Fellow at the School of Electrical Engineering, the University of Sydney, Australia. From June 2015 to now, he is a Professor at the School of Electronic and Optical Engineering, Nanjing University of Science and Technology, Nanjing, China. He was a visiting professor at Princeton University from 2018 to 2019. His research interests include network information theory, game theory, distributed intelligence, multiple agent reinforcement learning, and their applications in ultra-dense wireless networks, mobile edge computing, network privacy and security, and industrial Internet of things. He has co-authored more than 200 papers in IEEE journals and conferences, and holds 1 US patents and more than 10 Chinese patents in these areas. He is serving as an editor of IEEE Transactions on Wireless Communication and TPC member for several flagship IEEE conferences.
\end{IEEEbiography}
\begin{IEEEbiography}[{\includegraphics[width=1in,height=1.25in,clip,keepaspectratio]{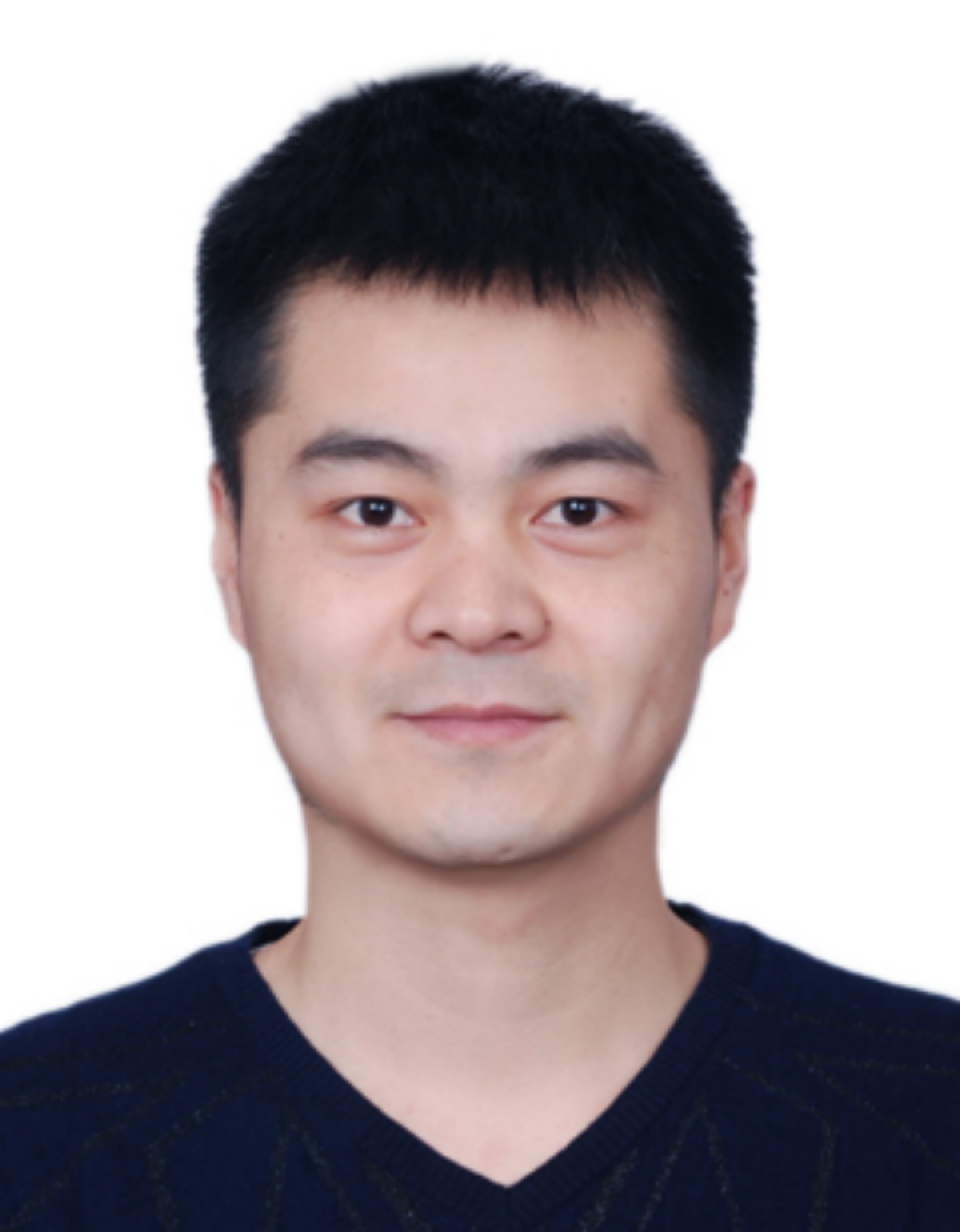}}]{Long Shi}
(Senior Member, IEEE) received the Ph.D. degree in Electrical Engineering from the University of New South Wales, Sydney, Australia, in 2012. From 2013 to 2016, he was a Postdoctoral Fellow at the Institute of Network Coding, Chinese University of Hong Kong, China. From 2014 to 2017, he was a Lecturer at Nanjing University of Aeronautics and Astronautics, Nanjing, China. From 2017 to 2020, he was a Research Fellow at the Singapore University of Technology and Design. Now he is a Professor at the School of Electronic and Optical Engineering, Nanjing University of Science and Technology, Nanjing, China. His research interests include blockchain networks, wireless communications, and federated learning.

\end{IEEEbiography}
\begin{IEEEbiography}[{\includegraphics[width=1in,height=1.25in,clip,keepaspectratio]{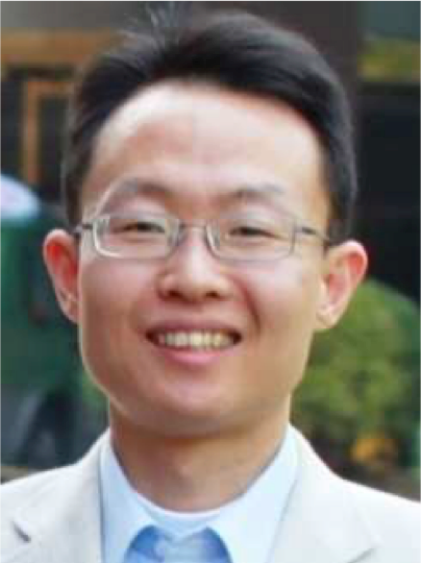}}]{Ming Ding}
(Senior Member, IEEE) received the B.S. and M.S. degrees (with first class Hons.) in electronics engineering from Shanghai Jiao Tong University (SJTU), Shanghai, China, and the Doctor of Philosophy (Ph.D.) degree in signal and information processing from SJTU, in 2004, 2007, and 2011, respectively. From April 2007 to September 2014, he worked at Sharp Laboratories of China in Shanghai, China as a Researcher/Senior Researcher/Principal Researcher. He also served as the Algorithm Design Director and Programming Director for a system-level simulator of future telecommunication networks in Sharp Laboratories of China for more than 7 years. Currently, he is a senior research scientist at Data61, CSIRO, in Sydney, NSW, Australia. His research interests include information technology, data privacy and security, machine Learning and AI, etc. He has authored over 100 papers in IEEE journals and conferences, all in recognized venues, and around 20 3GPP standardization contributions, as well as a Springer book ``Multi-point Cooperative Communication Systems: Theory and Applications”. Also, he holds 21 US patents and co-invented another 100+ patents on 4G/5G technologies in CN, JP, KR, EU, etc. Currently, he is an editor of IEEE Transactions on Wireless Communications and IEEE Wireless Communications Letters. Besides, he is or has been Guest Editor/Co-Chair/Co-Tutor/TPC member of several IEEE top-tier journals/conferences, e.g., the IEEE Journal on Selected Areas in Communications, the IEEE Communications Magazine, and the IEEE Globecom Workshops, etc. He was the lead speaker of the industrial presentation on unmanned aerial vehicles in IEEE Globecom 2017, which was awarded as the Most Attended Industry Program in the conference. Also, he was awarded in 2017 as the Exemplary Reviewer for IEEE Transactions on Wireless Communications.
\end{IEEEbiography}
\begin{IEEEbiography}[{\includegraphics[width=1in,height=1.25in,clip,keepaspectratio]{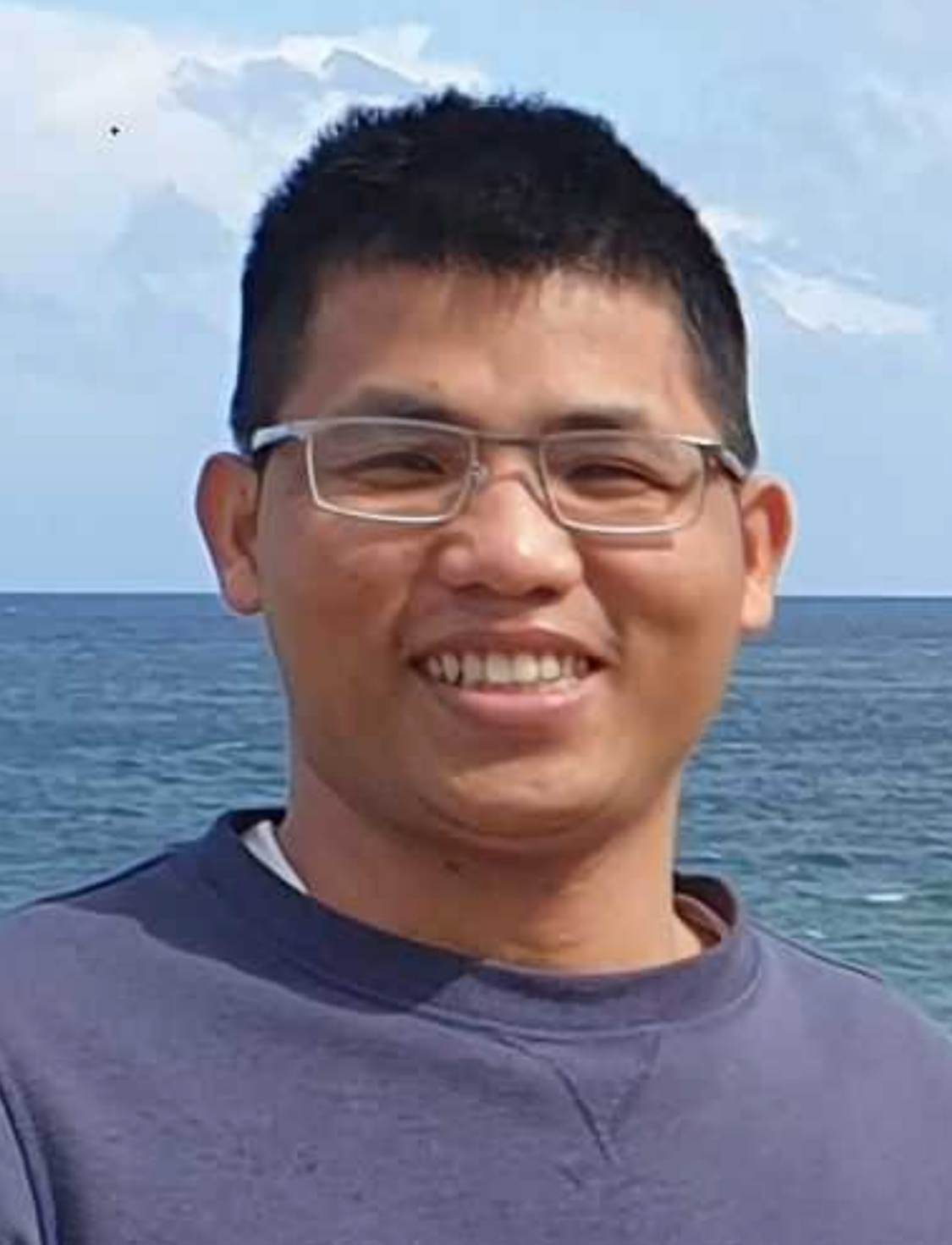}}]{Dinh C. Nguyen}
(Member, IEEE) is currently a postdoctoral researcher at the School of Electrical and Computer Engineering, Purdue University, USA. He received the Ph.D. degree in computer science from Deakin University, Australia in January 2022. His research interests include federated learning, blockchain, machine learning, and wireless networking. He has published over 20 first-author papers on top-tier IEEE/ACM journals and conferences such as EEE Journal on Selected Areas in Communications, IEEE Transactions on Mobile Computing, IEEE Internet of Things Journal, IEEE GLOBECOM and ICC conferences. He is an Associate Editor of the IEEE Open Journal of the Communications Society. He has received several research awards, including Web of Science ESI Highly Cited Paper Award.
\end{IEEEbiography}
\begin{IEEEbiographynophoto}{Wuzheng Tan}
received the B.S. degree from East China Jiao Tong University, Nanchang, China, in 1999, the M.S. degree from Guang Xi University, Nanning, China, in 2004, and the Ph.D. degree from the School of Computer Science and Technology, Shanghai Jiao Tong University, Shanghai, China, in 2008.
He is currently a Full Professor with Jinan University, Guangzhou, China. His research interests include security in digital assets, security in industrial control, and security in agricultural Internet of Things.
\end{IEEEbiographynophoto}
\begin{IEEEbiographynophoto}{Jian Weng}
(Member, IEEE) received the B.E. and M.E. degrees from the School of Computer Science and Technology, South China University of Technology, Guangzhou, China, in 2000 and 2004, respectively, and the Ph.D. degree from the School of Computer Science and Technology, Shanghai Jiao Tong University, Shanghai, China, in 2008.
He is currently a Full Professor with the College of Information Science and Technology, Jinan University, Guangzhou. His research interests include cryptography, information security, and artificial intelligence.
\end{IEEEbiographynophoto}
\begin{IEEEbiography}[{\includegraphics[width=1in,height=1.25in,clip,keepaspectratio]{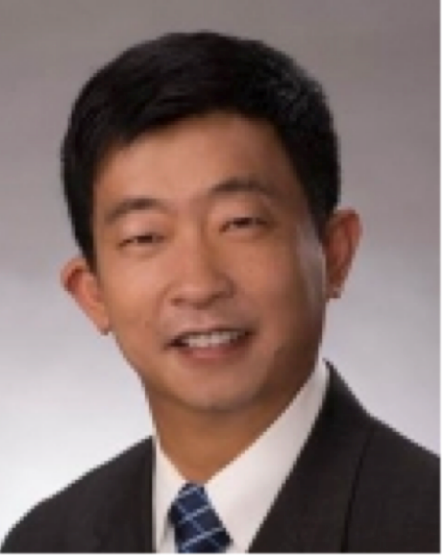}}]{Zhu Han}
(Fellow, IEEE) received the B.S. degree in electronic engineering from Tsinghua University, in 1997, and the M.S. and Ph.D. degrees in electrical and computer engineering from the University of Maryland, College Park, in 1999 and 2003, respectively. From 2000 to 2002, he was a Research and Development Engineer with JDSU, Germantown, Maryland. From 2003 to 2006, he was a Research Associate with the University of Maryland. From 2006 to 2008, he was an Assistant Professor with Boise State University, Idaho. He is currently a John and Rebecca Moores Professor with the Electrical and Computer Engineering Department and the Computer Science Department, University of Houston, Texas. He is also a Chair Professor with National Chiao Tung University, China. His research interests include wireless resource allocation and management, wireless communications and networking, game theory, big data analysis, security, and smart grid. He received the NSF Career Award, in 2010, the Fred W. Ellersick Prize of the IEEE Communication Society, in 2011, the Best Paper Award of EURASIP Journal on Advances in Signal Processing, in 2015, the IEEE Leonard G. Abraham Prize in the field of communications systems (best paper award in the IEEE JSAC), in 2016, and several best paper awards in the IEEE conferences. He has served as the IEEE Communications Society Distinguished Lecturer, from 2015 to 2018.

\end{IEEEbiography}

\end{document}